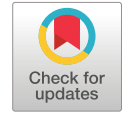

# Evaluation of State-of-the-Art Deep Learning Techniques for Plant Disease and Pest Detection

MD Tausif Mallick[1], Saptarshi Banerjee[2], Nityananda Thakur[3], Himadri Nath Saha[4,*] and Amlan Chakrabarti[1]

[1]Department of AKCSIT, University of Calcutta, Kolkata, 700106, India
[2]Department of Computer Science, Illinois Institute of Technology, Chicago, IL 60616, USA
[3]Department of Mathematics, SNEC, University of Calcutta, Kolkata, 700009, India
[4]Department of Computer Science, SNEC, University of Calcutta, Kolkata, 700009, India
*Corresponding Author: Himadri Nath Saha. Email: cshns@snec.edu.in

**ABSTRACT:** Addressing plant diseases and pests is not just crucial; it's a matter of utmost importance for enhancing crop production and preventing economic losses. Recent advancements in artificial intelligence, machine learning, and deep learning have revolutionised the precision and efficiency of this process, surpassing the limitations of manual identification. This study comprehensively reviews modern computer-based techniques, including recent advances in artificial intelligence, for detecting diseases and pests through images. This paper uniquely categorises methodologies into hyperspectral imaging, non-visualisation techniques, visualisation approaches, modified deep learning architectures, and transformer models, helping researchers gain detailed, insightful understandings. The exhaustive survey of recent works and comparative studies in this domain guides researchers in selecting appropriate and advanced state-of-the-art methods for plant disease and pest detection. Additionally, this paper highlights the consistent superiority of modern AI-based approaches, which often outperform older image analysis methods in terms of speed and accuracy. Further, this survey focuses on the efficiency of vision transformers against well-known deep learning architectures like MobileNetV3, which shows that Hierarchical Vision Transformer (HvT) can achieve accuracy upwards of 99.3% in plant disease detection. The study concludes by addressing the challenges of designing the systems, proposing potential solutions, and outlining directions for future research in this field.

**KEYWORDS:** Image processing; machine learning; deep learning; vision transformer; pest and disease detection

## 1 Introduction

In 2050, it's predicted that there will be 9.1 billion people on the planet [1] and, therefore, to supply the food needs of the rapidly expanding population, the agricultural output must increase by around 70%. Food security, rural livelihoods, and agricultural trade are all negatively impacted by plant pests and diseases, which cause up to 40% of worldwide crop production to be lost each year, according to the Food and Agriculture Organization (FAO) [2]. To meet the growing need for food, agricultural productivity must rise by almost 70%, as the world's population is expected to reach 9.1 billion by 2050 [1]. Early disease and pest identification and control are agronomic requirements and vital facilitators of food system resilience. Global food security goals are directly supported by emerging technologies like artificial intelligence and deep learning, which provide scalable and precise solutions that enable prompt intervention, less chemical use, and sustainable agriculture methods [2–6].

Fungal or fungal-like organisms primarily cause most plant diseases (around 85%). The bulk of plant diseases are generally linked to bacterial or viral pathogens, whereas only a small number are caused by certain worms [7]. According to recent research, pathogen-induced stress significantly alters chloroplast function and reduces photosynthetic efficiency, resulting in slower plant development and lower potential yield [8]. Among the most damaging fungal threats to the world's food security are wheat rust diseases, which include stem rust *("Puccinia graminis")*, stripe rust *("P. striiformis")*, and leaf rust *("P. triticina")*. Stripe rust alone can result in total crop failure under vulnerable circumstances, and several diseases can cause catastrophic crop losses, with yearly economic costs estimated at US $5 billion worldwide [9]. Pests also cause considerable losses in agricultural production [4] as insects often feed on leaves, disrupting photosynthesis. Furthermore, pests serve as vectors for various diseases, contributing to significant crop losses worldwide [3]. For example, the bug *"L.hesperus"* is a pest that severely damages cotton crops, and therefore, it is a common target of pesticide applications [10,11].

"Smart farming" refers to the management of farms that make use of contemporary technology for communication and information to maximize production and quality while minimizing the need for manpower [12]; it is crucial for raising agricultural production's yield, ecological impact, food safety, and resilience [13]. Smart farming enables the continuous monitoring of plants to detect diseases and pests at an early stage, allowing for timely interventions. When an infection is detected, smart farming can help farmers classify different types of diseases and pests based on various aspects, such as other plant and crop species, the age of plants, and external conditions [14–16]. Based on the identified threats, smart farming can be used to recommend appropriate countermeasures, such as pesticides, fungicides, or herbicides, to mitigate these risks. As a result, farmers can act very early and stop the spread of diseases and pests before they can infect large areas of the fields. Additionally, the use of toxic chemicals can be significantly reduced. For these reasons, fast, automated, accurate, and cost-effective approaches to identifying crop diseases and pests are of significant practical importance [17,18].

The extraction and analysis of plant image data have enabled several novel approaches, such as computer vision, machine learning (ML), and deep learning (DL), to successfully identify plant diseases and pests. These techniques are also accurate and fast. Rehman et al. secured an accuracy of 99.6% and above on apple, grape, peach and cherry diseases using Darknet53 while proposing an improved swarm optimisation butterfly algorithm [19]. Khan et al. [20] achieved an accuracy 97.2% with their proposed feature extraction method in detecting apple diseases. *Hassan et al.* have proposed a novel deep-learning model based on the inception layer, and a residual connection has been trained and tested on three different plant disease datasets, obtaining 99.39% accuracy on the PlantVillage dataset overall, and 99.66% on the rice disease dataset and the cassava dataset 76.59% [21]. The symptoms of pest attack or disease infection on plants differ in shape, texture, colour, and size based on the nature of pathogens, viruses, fungi, etc. The qualities of the image surface features, such as texture, colours, and structure, are assessed using deep learning [22], A branch of artificial intelligence called machine learning is being used more and more to mimic human cognitive processes like perception, categorization, and judgment in intricate, data-rich settings [23]. To be more precise, convolutional neural networks (CNN) are frequently used in various object identification and image classification applications [24–27]. These algorithms can be trained using various contaminated plant images gathered from public online sources, agricultural farms, and laboratories. Several image processing-based segmentation, feature extraction and clustering methods are applied to an input image for successful and accurate disease and pest detection. Some of the CNN models used in this domain include AlexNet [28], InceptionV3 [29], Inception V4 [30], VGGnet [31], Microsoft ResNet [32], and DenseNet [33].

Our review paper uniquely categorises and evaluates methodologies for identifying plant pests and diseases using deep learning models. Unlike existing literature, which often consolidates various methods into a single section, this structured categorization allows researchers to efficiently identify and compare the most suitable deep learning techniques for their specific pest and disease detection objectives. Additionally, we examine modified deep learning and transformer models, incorporating the latest research to reflect current advancements in the field. We have used vision transformers against well-established models on the tomato leaf dataset from PlantVillage [34]; the results demonstrated that vision transformers perform better than the convolutional neural network (CNN) architectures. This comprehensive analysis provides valuable insights, guiding researchers in selecting the most suitable methods for their needs and advancing the field. A general flow diagram of implementing deep-learning models for visualizing plant pests is described in Fig. 1, which states that the process begins with dataset acquisition, followed by its division into two halves. Consequently, the training/validation graphs highlight the models' significance, emphasizing the training methods of deep learning architectures, whether initiated from scratch or utilizing transfer learning. The following stage involves using assessment metrics specifically tailored for image classification tasks, particularly for identifying pest and plant disease varieties. The procedure involves applying mapping and visualization strategies that facilitate image classification, localization, and identification. This comprehensive visual depiction clearly illustrates the deep learning model's sequential workflow, providing readers with a clear understanding of its structure, implementation procedures, and general operation.

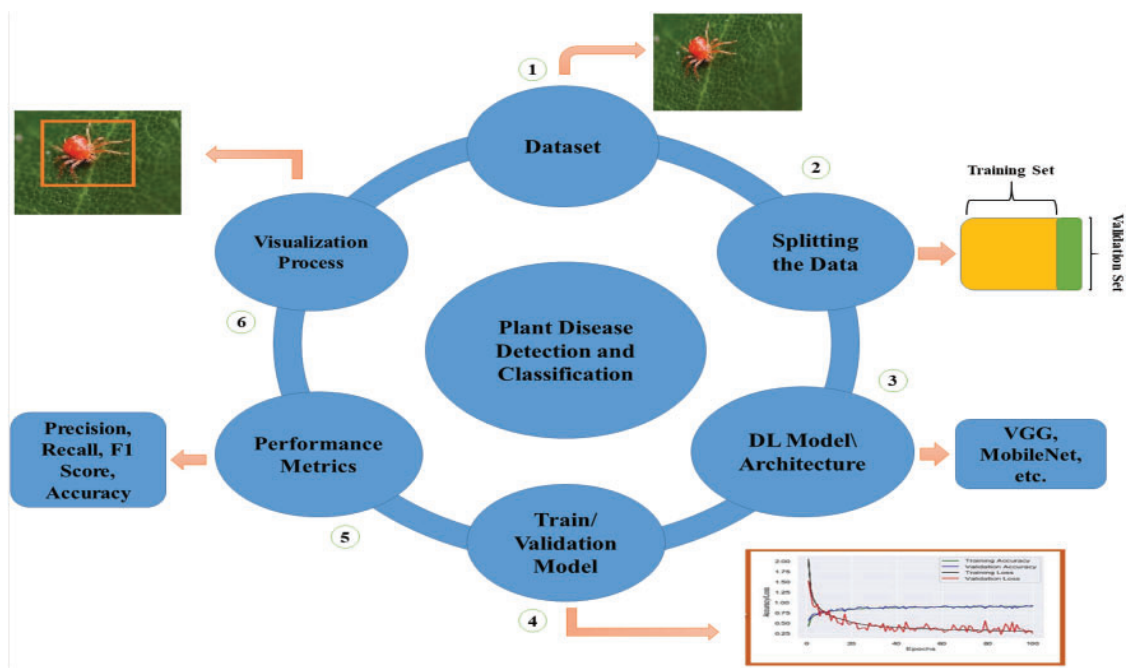

**Figure 1:** A comprehensive flow diagram illustrating the developmental stages of a typical deep learning model

This survey work offers several contributions to plant pest and disease detection:

- Different types of traditional and modified deep learning architectures for identifying and classifying pests and diseases of plants are described.
- CNN models for plant pest and disease detection have been compared. Researchers must select the appropriate deep-learning model to construct a realistic scenario for plant pest and disease identification in the real world.

- Visualization methodologies show the localization of infected spots on the plants and provide the farmers with necessary prevention suggestions about the pest or disease. Moreover, this agricultural information is presented without the involvement of agricultural professionals.
- An overview of hyperspectral imaging with deep learning methodologies is provided, along with its applications in both laboratory and field settings for classifying and identifying the early stages of pest attacks and disease lesions in plants.
- The paper consolidates the advantages and limitations of the techniques above, providing users with a comprehensive understanding to inform their selection of the most appropriate methodology for their specific applications.
- The architecture and suitability of the Vision Transformer (ViT) models for visual identification tasks are highlighted in this manuscript [35]. The performance of transformer-based architectures Swin Transformer [36], HvT [37], PMVT [38], Tiny-LeViT [39], and ViTAE [40] on the Tomato Leaf dataset from the PlantVillage dataset [34]. The results demonstrated that the vision transformers performed on par with and surpassed machine learning and CNN models in test and validation accuracy.
- Analyzes machine learning, deep learning, and vision transformers quantitatively in terms of precision, computing efficiency, and visualization capabilities.
- The scope of future research on pest and disease identification in agriculture, including potential improvements, general limitations and challenges, has been aggregated.

## 2 Structural Design of the Survey Work

The bibliography-based analysis of plant pest and disease detection and classification involves two steps in the study: (a) accumulation of existing related research works and (b) a comprehensive evaluation of various research works. The journal articles and conference papers were collected from the scientific knowledge bases of IEEE Explore [41] and ScienceDirect [42]. As search keywords, the following queries are applied: "agriculture", "deep learning", "plant pest and disease detection," and "convolutional neural network."

We have sorted out papers that describe traditional and Modified deep learning models applied to the pest and disease domain of the agricultural field. We initially collected over a hundred papers covering all the methodologies relevant to this work. In the next phase, the selected papers were evaluated individually, focusing on the different characteristics, such as accuracy and time, as well as the working principles of these technologies.

This is a structural breakdown of this paper:

- Section 3 outlines the typical deep learning model visualization and non-visualization techniques, other modified architectures, and state-of-the-art transformer models for identifying plant pests and diseases.
- Section 4 explains hyperspectral imaging with deep learning models.
- Section 5 offers the research challenges and possible solutions in this domain.
- Section 7 provides a nuanced comparison between traditional machine learning, CNNs and contemporary vision transformers.
- This paper has been concluded in Section 8 and provided future recommendations to improve the visualization and identification of diseases and pests of plants.

In Fig. 2, the organization of the paper is presented.

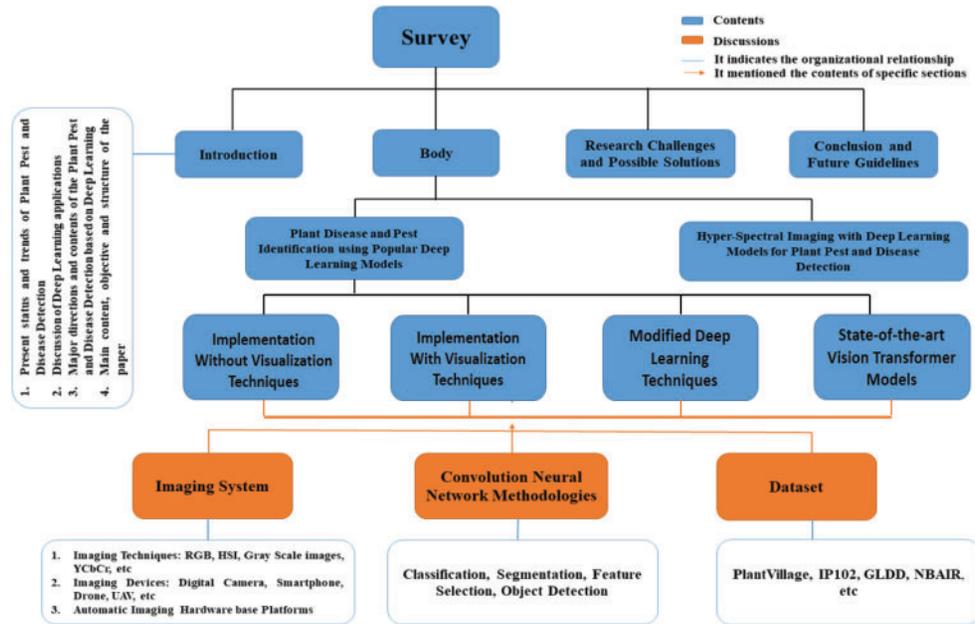

**Figure 2:** The organisation of the paper

## 3 Detection of Plant Disease and Pests Employing Popular Deep Learning Models

The concept of deep learning was created by Hinton et al. in 2006 [43]. Table 1 presents the differences between conventional image processing and machine learning approaches with deep learning methodologies. Many traditional deep learning architectures have been developed since the introduction of AlexNet [28]. For instance, Patel et al. demonstrated that state-of-the-art EfficientNet architectures achieve 98.3% disease classification accuracy while requiring 8.4 times fewer parameters than conventional CNNs [44].

**Table 1:** Comparison of deep learning approaches and traditional image processing techniques

| Capabilities | Conventional image processing techniques | Deep learning methodologies |
| --- | --- | --- |
| Principle | Manual feature design followed by classification. | Automated feature extraction and learning from large datasets [45]. |
| Technique | Image segmentation methods: thresholding, edge detection, and region-based segmentation. Feature extraction: SIFT [46], HOG [47], LBP [48], texture, structure, and color analysis. Classification methods: SVM [49], BP [50], Bayesian [51], etc. | Convolutional Neural Networks (CNNs) [24]. |

(Continued)

**Table 1 (continued)**

| Capabilities | Conventional image processing techniques | Deep learning methodologies |
|---|---|---|
| Necessary conditions | Requires controlled image capture, clear contrast between healthy and diseased areas, and low noise environments [15]. | Requires large labeled datasets and high computational resources, e.g., VGG (138 million parameters) [52]. |
| Appropriate situations | Due to fluctuating conditions, performance frequently deteriorates in dynamic situations, necessitating frequent algorithm tuning (such as threshold adjustments) to preserve accuracy [15,45]. | Robust to complex and variable natural environments [39]. |

Many traditional deep learning architectures have been developed since the introduction of AlexNet [28]. For instance, Patel et al. demonstrated that state-of-the-art EfficientNet architectures achieve 98.3% disease classification accuracy while requiring 8.4 times fewer parameters than conventional CNNs [44]. Deep learning techniques are employed to identify, segment, and categorise images. This section covers the well-known deep learning model-based research efforts used for plant disease and pest identification and classification in the agricultural domain. Notably, Lee and Wang recently proved that few transformer-based models can outperform CNNs in cross-species pest identification using attention mechanisms [53]. Additionally, various performance metrics are applied to these works to evaluate the designated deep learning models, as illustrated in Section 3.1.

### 3.1 Implementation of Deep Learning Models

The decision-making processes employed by deep learning architectures must be fully understood. Based on the properties of plant images, we synthesize two study approaches to extract insights. First, we have discussed the deep learning architectures based on (with or without) visualisation techniques. Next, the modified deep-learning architectures for plant pests and disease identification have been presented. In Table 2, the categories of deep learning models frequently employed to recognise and categorise plant diseases and pests are briefly explained.

#### 3.1.1 Without Visualization Technique

The classification of diseases affecting the maize plants was performed using a CNN, and the model's usefulness was determined using histogram approaches [54].

**Table 2:** The comparison of traditional deep learning architectures

| Deep learning architectures | Number of parameters | Important characteristics with advantages/disadvantages |
|---|---|---|
| AlexNet [28] | 60 M | Introduced ReLU and dropout; significant boost in image recognition. |
| VGG [52] | 138 M | Deep CNN using 3 × 3 filters; high computational cost due to many parameters. |
| LeNet [55] | 60 k | One of the earliest CNNs; minimal parameters, limited to basic computations. |

**Table 2 (continued)**

| Deep learning architectures | Number of parameters | Important characteristics with advantages/disadvantages |
|---|---|---|
| OverFeat [56] | 145 M | Unified CNN for classification, localization, and detection; more complex than AlexNet. |
| ZFNet [57] | 42.6 M | Enhanced AlexNet with smaller filter and stride sizes; improved accuracy. |
| GoogleNet [58] | 8 M | Fewer parameters than AlexNet; strong accuracy using Inception modules. |
| ResNet [59] | 23 M | Introduced residual connections to address vanishing gradients; superior accuracy. |
| DenseNet [60] | 7.1 M | Uses dense connections; compact yet effective with high performance. |
| SqueezeNet [61] | 1.25 M | Tiny model with $1 \times 1$ filters; fast and efficient, comparable to AlexNet. |
| Xception [62] | 22.8 M | Employs depthwise separable convolutions; outperforms VGG, ResNet, and Inception-v3. |
| MobileNet [63] | 13 M | Lightweight model using depthwise separable convolutions; near VGG-level accuracy. |
| Modified/Reduced MobileNet [64] | 0.6/0.55 M | Fewer parameters than original MobileNet; matches its accuracy. |
| VGG-Inception [65] | 134 M | Combines VGG and Inception; replaces $5 \times 5$ conv with two $3 \times 3$ layers; improves test accuracy over popular architectures. |

Note: k = kilo or thousand, M = Million

The extraction of pertinent visual features made possible by this technique significantly improved the categorization accuracy. Additionally, the model showed consistent performance across a range of imaging circumstances, indicating that it is robust for field-level applications. Tomato plants are highly vulnerable to a variety of diseases that can negatively affect their quality and yield. To detect various tomato plant diseases, the standard CNN architectures "AlexNet," "ResNet," and "GoogleNet" were introduced in [66]. According to the experimental results, the "ResNet" performed better than the other convolutional neural network models applied. The "LeNet" Architecture identified the banana leaf diseases, which significantly supported an accurate detection of RGB images with 92.88% detection accuracy and 85.94% for Grey Scale images of banana diseases in [67]. Different types of deep learning models, including "VGG", "AlexNet", "AlexNetOWTbn", "GoogLeNet", and "Overfeat", were applied for the recognition of plant diseases and lesions [68]. Eight different plant diseases are identified using a variety of machine learning classifiers, including the following: "SVM", "KNN", and "Extreme Learning Machine (ELM)," as well as deep learning architectures such as "GoogleNet," "ResNet-50," "InceptionResNetv2," and "SqueezeNet" [69]. Finally, the "ResNet-50" with the "Support Vector Machine" classifier outperformed other trained models.

In [70], three types of diseases and two pest infections of the cassava leaf dataset were identified by applying transfer learning of the "Inception-v3" architecture. A proprietary convolutional neural network and a 4-fold cross-validation approach were used to construct an autonomous and reliable plant disease identification model, and the emerging algorithm had an average precision of 82.3% [71]. The deep learning classifier categorised four potato plant diseases in [72]. The convolutional neural network models employed were VGG16 and VGG19. Finally, the VGG16 model obtained an average accuracy of 91% higher than the accuracy of 90% (average) achieved by the VGG19 model for potato plant disease classification. Four different tomato plant diseases were successfully identified and categorised using a hybrid model of CNN and the "Learning Vector Quantization (LVQ)" technique, with an average classification accuracy of 86% in [73]. In [74], a comparative study was presented to select the most suitable deep learning architecture for identifying plant diseases. Furthermore, in [75], the PlantVillage dataset was utilised for training the deep learning-based architectures "VGG16" and "AlexNet," which both attained classification accuracy of 97.29% and 97.49%, respectively, using the healthy class and six-leaf disease classes of tomato plants. In the methodologies mentioned above, Grad-CAM and Score-CAM visualisation methods are not used to spot infections of plant diseases and pest lesions. Table 3 summarises the key advantages and limitations of using non-visualisation techniques in deep learning models for plant disease and pest detection.

**Table 3:** Advantages and limitations of non-visualization techniques for plant disease and pest detection

| Advantages | Disadvantages |
| --- | --- |
| Deep learning models AlexNet [28], VGG [52], and ResNet [59] achieve high accuracy in disease and pest classification, even without visualization. Notably, VGG [52] and ResNet [59] have shown exceptional performance. | Visualization is essential for understanding model decision-making; its absence complicates validation, refinement, and trust in model robustness. |
| Non-visual techniques leverage computational efficiency of deep learning models, allowing rapid processing and classification of large datasets. | These techniques run the risk of missing small but important clues because they may mask important traits or patterns required for precise detection. |
| Simplifies the workflow by focusing only on the model's outputs, avoiding the complexity of interpreting visual data. | The lack of interpretability makes it difficult to validate or explain predictions, reducing model transparency and acceptance in agricultural practice. |
| Boosts productivity, minimizes human error, and supports automation in plant disease management. | Heavy reliance on preprocessing for high accuracy can be resource-intensive and may miss fine-grained information critical to accurate detection. |

### 3.1.2 With Visualization Technique

Plant pest and disease symptoms have been better visualized and understood using several deep learning architectures. Saliency maps [76] and occlusion approaches [77] have drawn attention to disease-affected areas, making them easier for farmers to understand. An Ultra-Lightweight Efficient Network (ULEN) outperformed SVM in [78] with a classification accuracy of 96.30%. Nine tomato leaf diseases were classified using ResNet-50, Xception, MobileNet, ShuffleNet, and DenseNet121_Xception in [79], with early convolution layers successfully identifying impacted areas. In [80], an enhanced LeNet used edge mapping to identify localized olive plant diseases. Reference [81] suggested a deep model for cucumber plants that accurately outperformed SVM, RF, and AlexNet. YOLOv4 produced the most favourable classification results when object detection models, such as SSD, YOLOv4, and Faster R-CNN, were tested for scale pest localization [82]. In [83], a novel teacher-student framework enabled region-level visualization and precise lesion identification.

The percentage of plant regions affected by disease has been accurately determined using deep learning models combined with different detectors [84]. Faster R-CNN, R-FCN, and SSD combinations with backbone networks including AlexNet, VGG, ZFNet, ResNet-50, and ResNetXt-101 were tested; the most efficient results were obtained with R-FCN + ResNet-50—the use of bounding boxes allowed for accurate disease localization. Using ResNet-50, Inception-V2, and MobileNet-V1 combined with Faster R-CNN and SSD, banana leaf diseases were identified in [85]. Similarly, a CNN-based pipeline for maize detected northern leaf blight (NLB) lesions, which were then evaluated using receiver operating characteristic (ROC) curves and annotated using CyVerse tools [67]. Accurate in-situ detection of rice leaf disease was made possible by Faster R-CNN with an improved Region Proposal Network (RPN) architecture, which produced high-quality regions of interest [86].

Many deep learning models have been successfully used to detect pests and plant diseases. VGG-based CNN and fully convolutional networks displayed wheat disorders [87], but LeNet correctly recognized soybean diseases [88]. In [89], radish regions were divided into healthy and Fusarium wilt-affected areas using the VGG model, with bacterial patches being identified using K-means clustering. Similar deep learning techniques were used to visualize plant symptoms in [90], and UNet and PSPNet with ResNet-50 were used to classify the severity and stress of coffee leaves based on disease spots [91]. Hidden neurons for pest detection were adjusted by an improved YOLOv3 using Adaptive Energy-based Harris Hawks Optimization (AE-HHO) [92]. Tiny YOLOv3 detected multi-class insect sites on sticky traps [93], while CNNs indicated areas of rice crops impacted by infection [94].

In order to guarantee color stability, target spots were identified from segmented images using a hotspot-based technique that uses descriptors for bacterial color characteristics and scorching spot textures [95]. In [96], field crop insects were classified using AlexNet, ResNet, GoogLeNet, and a suggested DCNN model; the DCNN demonstrated good reliability. To efficiently identify and categorize multi-class pests, "Pest-Net," a region-based end-to-end architecture first presented in [97], included a Channel-Spatial Attention (CSA) unit, Region Proposal Network (RPN), Fully Connected (FC) layers, and Position-Sensitive Score Map (PSSM). A hybrid model that combined ZFNet with RPN and Non-Maximum Suppression (NMS) performed better than more conventional models like AlexNet and ResNet for the localization and counting of insects [98]. In [99], fruit crop diseases were automatically segmented and recognized using visualization techniques based on VGG-16 and AlexNet. The real-time model "Faster DR-IACNN" successfully identified grape leaf diseases in [100], whereas LeNet, in conjunction with color space and vegetative indicators, successfully identified grape plant diseases in [101].

By combining dilated convolution and global pooling, the Global Pooling Dilated Convolutional Neural Network (GPDCNN) successfully recognized six prevalent cucumber leaf diseases in a different study [102].

In situ experimental observations must consider real-world circumstances to assess how well deep learning architectures identify plant diseases and pests. Their application for practical disease detection and classification is limited because most of the approaches covered above rely on datasets gathered in controlled or typical environmental circumstances [50,65–67,77–80,94]. Real-time field circumstances have been addressed explicitly in only a small number of studies [66,69,70,84,87,89,102]. Several of these works' visualization techniques are depicted in Figs. 3–11.

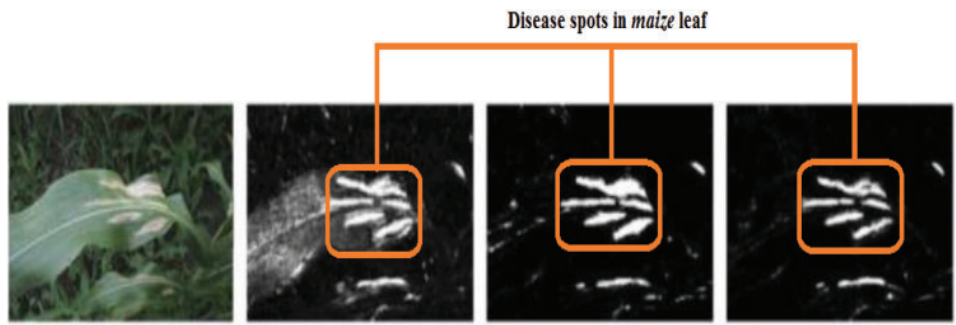

**Figure 3:** Identification of maize disease (specified by orange rectangles) by heat map [45]

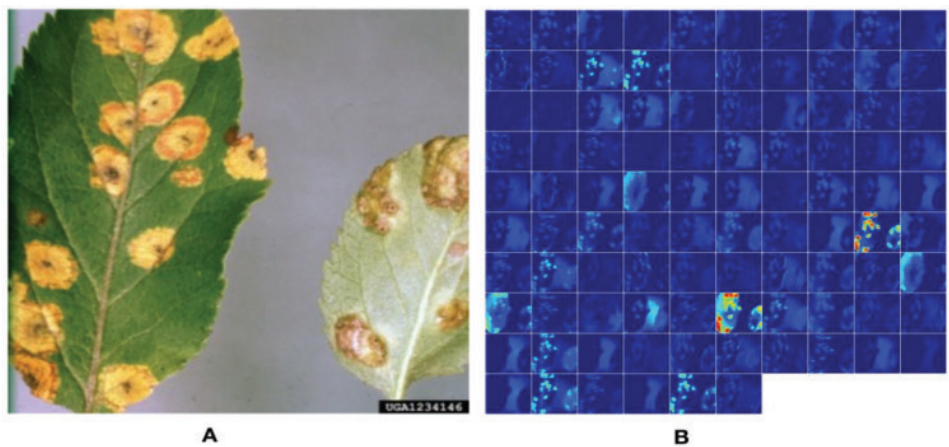

**Figure 4:** Demonstration of activation in the first layers of an AlexNet model to show that it has learned to efficiently stimulate against the affected areas of the target leaf image of "Apple Cedar Rust" (a deep blue circle indicates the diseased patches) [79]

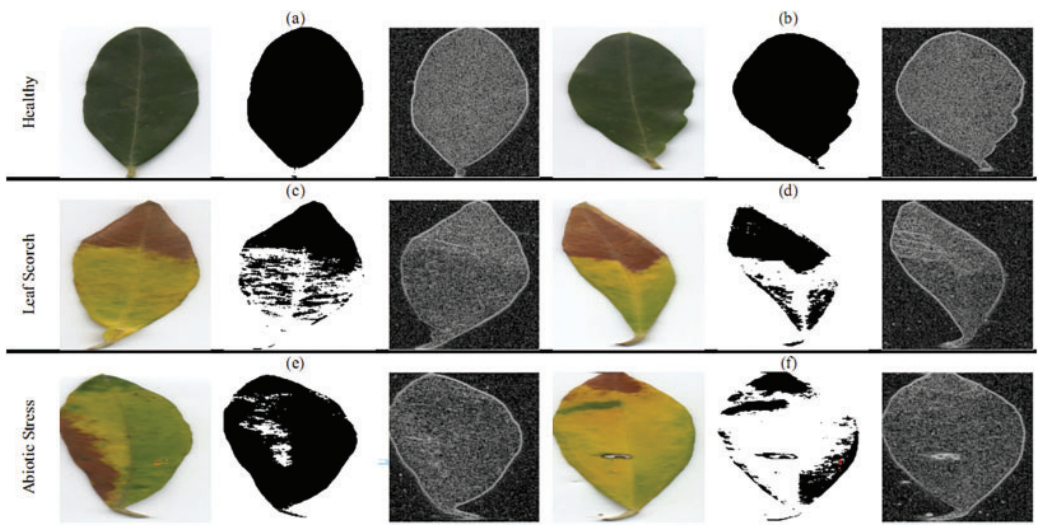

**Figure 5:** Olive leaf disease identification using segmentation and edge map [80]

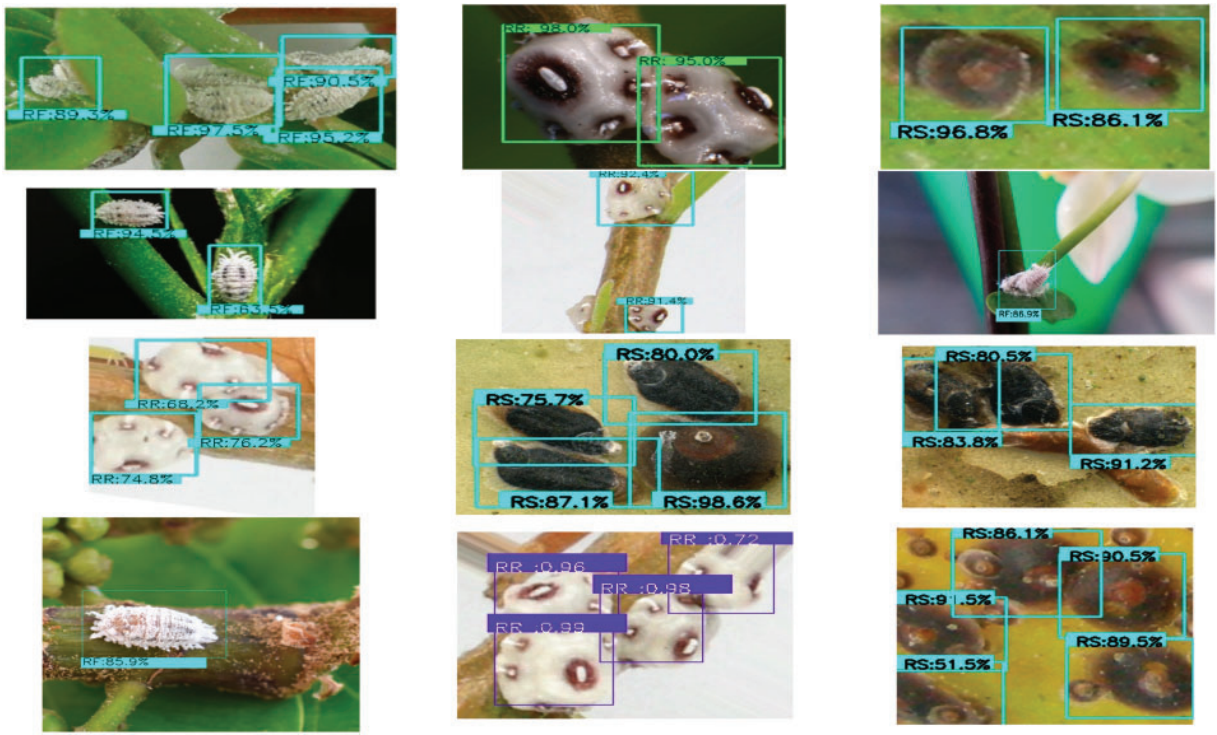

**Figure 6:** Bounding boxes mark the predicted outcomes of the object detection model to identify three specific pests [82]

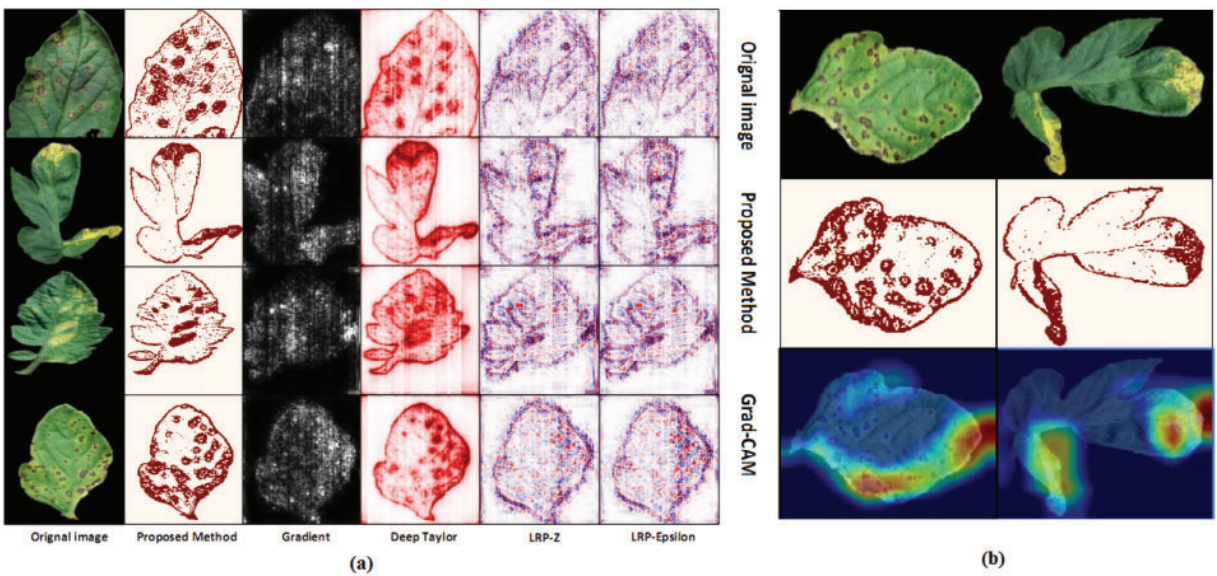

**Figure 7:** (**a**) Visualization algorithm results are compared; (**b**) Comparison of Grad-CAM [82]

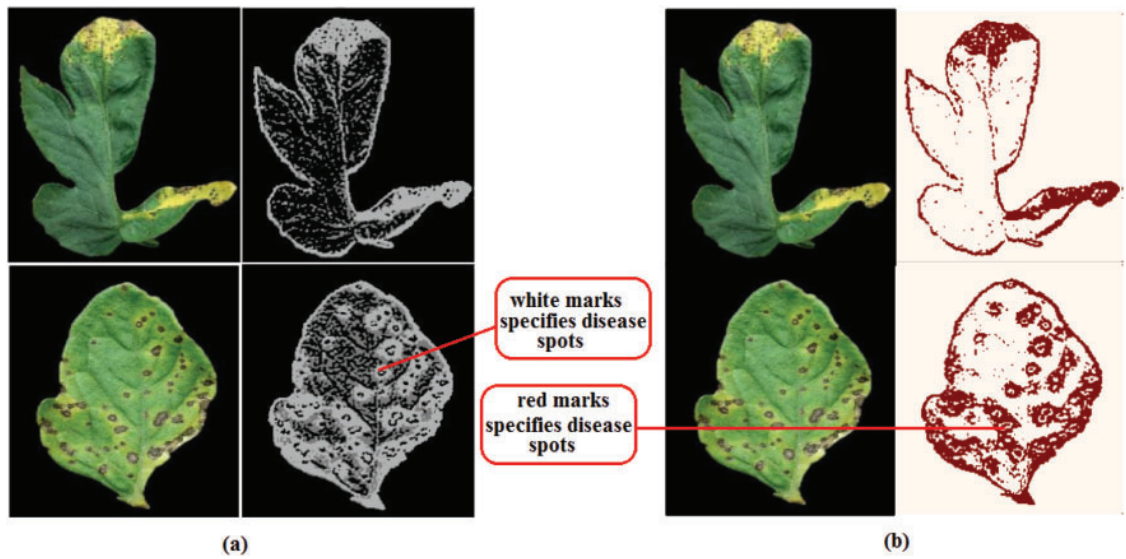

**Figure 8:** (**a**) A novel "*Teacher/student*" framework; (**b**) "Binary threshold algorithm" for fragmentation [83]

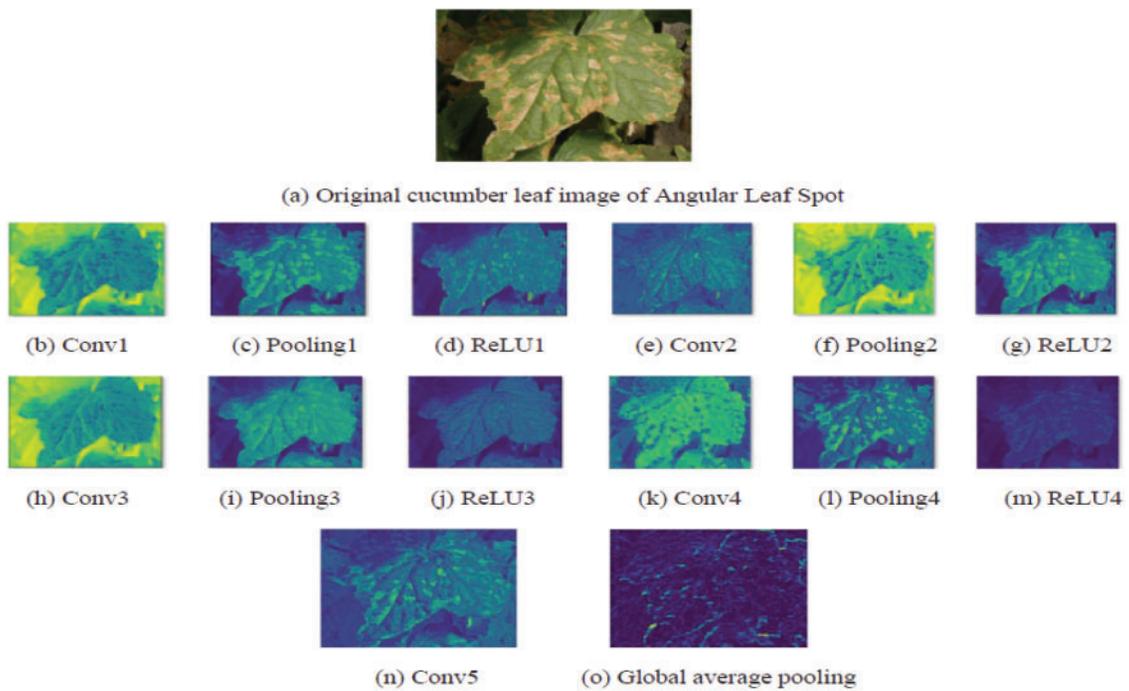

**Figure 9:** Feature maps representation after the convolution layers applied to an image [102]

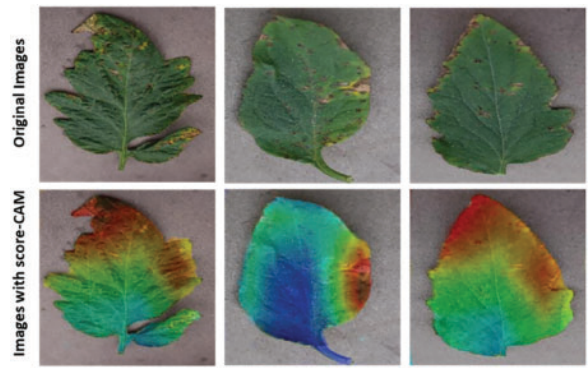

**Figure 10:** Visualization of tomato leaf pictures using Score-CAM with accurate classification: sliced leaves with a score-CAM heat map over the original leaves [103]

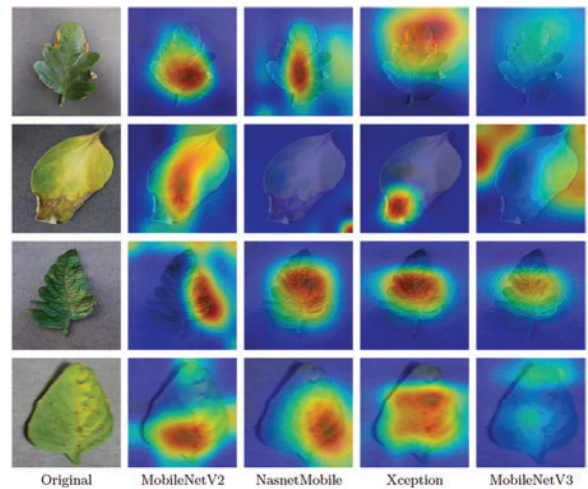

**Figure 11:** Saliency maps for "Bacterial spot," "Late blight," "Leaf mold," and "Yellow leaf curl virus" are shown in the first, second, third, and fourth rows, respectively [104]

Fig. 3 identifies the infected maize plants using the heat map. Initially, each location of the selected image contains a lesion, which can be considered a measure of the probability that each lesion was caused by the disease. After that, a matrix form is generated using the probabilities, indicating the model's outcome for each target image region [45]. The feature maps in Fig. 4 were obtained to understand the importance of CNN topologies in discriminating between different plant diseases. The output demonstrates the decent efficiency of the "AlexNet" architecture, which is capable of detecting plant diseases [79]. This figure also illustrates how characteristics are extracted using shallower Convolutional Neural Networks (CNNs). Compared to deeper models such as ResNet, the coarse and less selective activations of AlexNet may hamper performance in fine-grained disease classification. In Fig. 5, the plant diseases are detected using segmentation and edge maps. In the segmentation map, the yellow region appears as a white-coloured surface to highlight the lesion portion of the input leaf [80]. Segmentation, as opposed to bounding boxes, offers distinct boundaries, enhancing the quality of training data and the prediction of disease severity. Separating clustered leaf symptoms is further aided by edge-enhanced segmentation. The matrix reveals the probabilistic occurrence of lesions across spatial areas. By combining probabilities over time, these heatmaps may assist in improving

 

severity assessment, allowing for customized treatment as opposed to uniform spraying. In Fig. 6, the three classes of pest detection are presented. RF points to the "*mealybugs*" pests, RR means "*Coccidae*", and RS indicates "*Diaspididae*". Each image contains multiple predicted outcomes [82], and for class-wise detection, including mealybug and coccidae, is depicted in the figure. Visually, the detection seems accurate, but no metrics are displayed. A detailed comparison in the updated publication demonstrates that SSD achieves 82.5% mAP on the dataset, with an accuracy of 89% for mealybugs and 76% for coccidae. illustrates how distillation sharpens the student model's focus, enhancing interpretability and mitigating a novel approach for visualisation is demonstrated in [83], as presented in Figs. 7a,b and 8a,b. The selected imagery has been reassembled in Fig. 8a to create the "student/teacher" model [83], and in Fig. 8b, a single channel heat map has been created and segment significant portions of the image. After that, a straightforward Boolean threshold method reliably displays signs of severe plant infection. This design blends a student's efficiency with high-level semantics from a teacher model. This image illustrates how distillation sharpens the student model's focus, enhancing interpretability and mitigating overfitting on real training data. Fig. 7a presents the different visualisation approaches. Heat maps created using visualisation approaches like "Gradient," "LRP-Z," and "LRP-Epsilon" are hazy and difficult to understand. The deep Taylor technique has achieved superior outcomes, but it has specified only a few spots of the disease on the leaf. Furthermore, a defective localisation of the disease spots is achieved using the Grad-CAM technique. At the same time, the proposed method addressed the issue by employing a decoder, as shown in Fig. 7b [82]. As a conclusion of the figure, Deep Taylor and LRP concentrate on relevance at the pixel level, whereas Grad-CAM emphasises more general regions. Although LRP provides superior noise reduction in the updated statement, it may overlook contextual cues. According to a user assessment, agronomy specialists rated Grad-CAM as the most interpretable. Fig. 9 shows how the feature map uses pooling, convolution, and ReLU operations to process an image so that different features can be extracted from other layers. Every feature map is depicted, and ultimately, the Global Average Pooling (GAP) layer records several noteworthy attributes [102]. Additionally, the progressive abstraction of characteristics in a CNN is illustrated in this figure. Deeper layers identify intricate disease patterns, such as lesions or pest forms, while early layers record low-level information, such as edges and textures. For reliable categorization in a variety of illumination and occlusion scenarios, this layered abstraction is essential. Figs. 10 and 11 illustrate two distinct visualization techniques: saliency maps and heat maps. The Score-Weighted Visual Explanations for CNNs (Score-CAM) heat map is used to identify infections in the source images of Fig. 10 as red dots [103] and saliency maps aid in seeing the active pixels affecting the model's prediction, but noisy backgrounds may make them difficult to understand. They are faster than CAM-based techniques, although they are frequently less accurate in pinpointing the locations of symptoms. A superior methodology, known as the saliency map technique, is introduced, which effectively detects all disease-affected spots on tomato plants, as shown in Fig. 11 [104]. This technique generates more concentrated attention maps than saliency, a class-specific activation. This figure illustrates how Score-CAM enhances field-level diagnostic accuracy by distinguishing between overlapping symptoms of disease. Table 4 shows different visualisation mapping/techniques to summarise this section. These methods, which highlight key image regions that make predictions, such as Grad-CAM, saliency maps, and feature attribution, improve the interpretability of deep learning models. In crucial applications like plant disease and pest detection, this kind of visualization aids in determining the model's target region, fostering trust and facilitating validation. Table 5 summarizes the key advantages and limitations of visualization techniques in deep learning models for plant disease and pest detection.

**Table 4:** Visualization mapping/approaches applied on different deep learning methods

| Visualization approaches/mappings | Reference |
| --- | --- |
| Heat maps were employed to locate the bacteria's hotspots | [45] |
| Display of activation | [65] |
| Feature diagram for the disease-affected rice plant | [75] |
| Illustration of a saliency map | [76] |
| Method for perceiving symptoms | [77] |
| Visualization of attributes sorted from the top to the bottom layer | [78] |
| Display the first convolutional layer's activations | [79] |
| Boundary map and segmented map | [80] |
| A mesh graph representation, a saliency map, and 2D and 3D outlines | [81] |
| Heat map construction, image segmentation by threshold theorem, and regeneration on distinctive areas | [83] |
| Disease location and categorization employing boundary boxes | [84] |
| Maps with spatial and characteristic metadata | [87] |
| Color space via K-means cluster and HSV | [89] |
| Position-Sensitive Score Map (PSSM) module | [97] |
| Methodology of image segmentation | [99] |
| Feature graph for identifying diseases | [102] |
| Representation of a saliency map | [104] |

**Table 5:** Advantages and limitations of visualization techniques for plant disease and pest detection

| Advantages | Disadvantages |
| --- | --- |
| Understanding decision-making by highlighting disease-affected regions; transparency aiding model validation and refinement [83]. | The effectiveness of visualization techniques hinges significantly on the quality and quantity of input data [54]. Inconsistent or low-quality data may result in misleading visualisations and inaccurate model interpretations. |
| Helps validate models by clarifying disease and pest classification processes, aiding precise model adjustments for improved accuracy and reliability [102]. | Although visualization techniques improve interpretability, the complexity of some methods can challenge accurate interpretation, particularly for non-expert users [79]. |
| Visualization techniques combined with deep learning models have enhanced classification accuracy and robustness, exemplified by GoogLeNet [58] and ResNet-50 [85]. | Models may overfit visual patterns in training data, compromising their ability to generalise to new, unseen data [86]. |
| Important metrics can be continuously monitored using data visualization tools to construct dashboards that update automatically [86]. | They do not always consider environmental variability, such as lighting changes or plant growth stages, which impact the accuracy of disease detection and localisation [45]. |

### 3.1.3 Modified Deep Learning Architectures for Plant Pests and Diseases Identification

Several new deep learning architectures have been applied in some works to achieve superior and apparent identification of plant diseases. For instance, reference [105] provided an improved "GoogLeNet" architecture, and this model's usefulness is significantly superior to that of the "AlexNet" and "VGG" architectures. The updated versions of these traditional deep learning models have achieved 98.9% accuracy. In [106], a fresh combination of "DenseNet" with "Xception" based "DCNN" is proposed for the recognition & early diagnosis of apple tree leaf diseases accurately. The proposed approach is 98.82 percent accurate in identifying five diseases in apple leaves. To identify, categorise, and count pests, a unique "entire-and-partial feature transfer learning" approach is presented; finally, it provides the detection frequency (90.2%) of pest movement [107]. Different fine-tuned feature maps are reinforced utilising the "entire-feature transfer learning" weight matrix while utilizing the "partial-feature transfer learning" approach. Finally, the whole feature network's cross-layer and multi-scale convolution layers are integrated. To categorise the soybean pests, transfer learning techniques are employed in [108] and in conjunction with the deep learning models "Inception-v3", "VGG-19", "Resnet-50", and "Xception". The entire soybean dataset comprises 5000 images, captured in real-time. The ResNet-50 deep learning model achieves the highest classification accuracy at 93.82%. Haridasan et al. [109] conducted a study focusing on the automatic detection and analysis of rice crop diseases by incorporating various pre-processing techniques and utilising Principal Component Analysis for dimensionality reduction. In their work, Abbas et al. [110] introduced an enhanced Cuckoo Search algorithm termed the Hybrid Improved Cuckoo Newton Raphson Optimization (HICNRO). This approach not only diminishes computational time but also eliminates redundant features and enhances accuracy. These selected features are then utilised to train ten distinct machine learning classifiers, resulting in notable accuracies of 100%, 92.9%, and 99.2% on Tomato, Potato, and Cucumber Leaf datasets. Khattak et al. [111] proposed a CNN model with a test accuracy of 94.55 %, making it a valuable decision support tool for farmers to classify citrus fruit/leaf diseases.

In [112], LeafNet, a new CNN model, is employed for tea leaf disease classification, achieving greater accuracy compared to the "Multi-Layer Perceptron (MLP)" and "Support Vector Machine (SVM)" models. In [113], a "deep convolutional generative adversarial network (DCGAN)" was trained on a segmented and filtered dataset containing 56,000 images of 19 crops, which comprise 38 classes collected from PlantVillage. A CNN model such as "MobileNet" deployed on smartphones is used to classify plant diseases and obtained 92% test accuracy compared to the modified "Inceptionv3" model with 88.6% accuracy. Two deep learning architectures, namely the reduced "MobileNet" and reformed "MobileNet", are proposed in [64], and the obtained classification accuracy is similar to that of the "VGG" architecture. Finally, the reduced "MobileNet" has obtained 98.34% accuracy. This model significantly reduces training time, as it has fewer parameters than the VGG model. In [65], the hybrid CNN architecture, known as the "VGG-Inception" model, is used to classify and identify five types of diseases in apple plants. This model outperformed several deep learning models, including AlexNet, GoogLeNet, VGG, and specific variants of ResNet. In addition, it offers inter-object and class identification and activation visualisation. Another advantage of using this model is that it perfectly visualises disease-affected spots on plants. In a different study, Liu et al. [114] developed an enhanced Faster R-CNN with feature pyramid fusion (FPF-RCNN) that maintained a detection speed of 21.5 FPS while achieving 79.2% mAP on the Citrus Pest and Disease Dataset (CPDD). Furthermore, Amin et al. [115] utilise two pre-trained convolutional neural networks (CNNs), EfficientNetB0 and DenseNet121, to extract deep features from the corn plant images, ultimately achieving a classification accuracy of 98.56%.

The most prominent deep learning algorithms for accurate plant infection identification and categorisation are shown in a bar chart in Fig. 12. Most research work has applied the AlexNet [28] model. The following widely used deep learning architectures are ResNet-50 [85], VGG-16 [75], and GoogLeNet [58].

Correspondingly, various hybrid deep learning architectures, such as VGG-Inception [65], reduced or modified MobileNet [64], updated LeNet [80], and modified GoogLeNet [105], are employed to identify pests and diseases in plants. In [116], Pacal et al. (2024) conducted a comprehensive review analyzing 160 studies on deep learning applications in plant disease detection, focusing on classification, detection, and segmentation tasks. The paper highlights the effectiveness of deep learning in early disease detection and discusses challenges such as dataset limitations and model generalization.

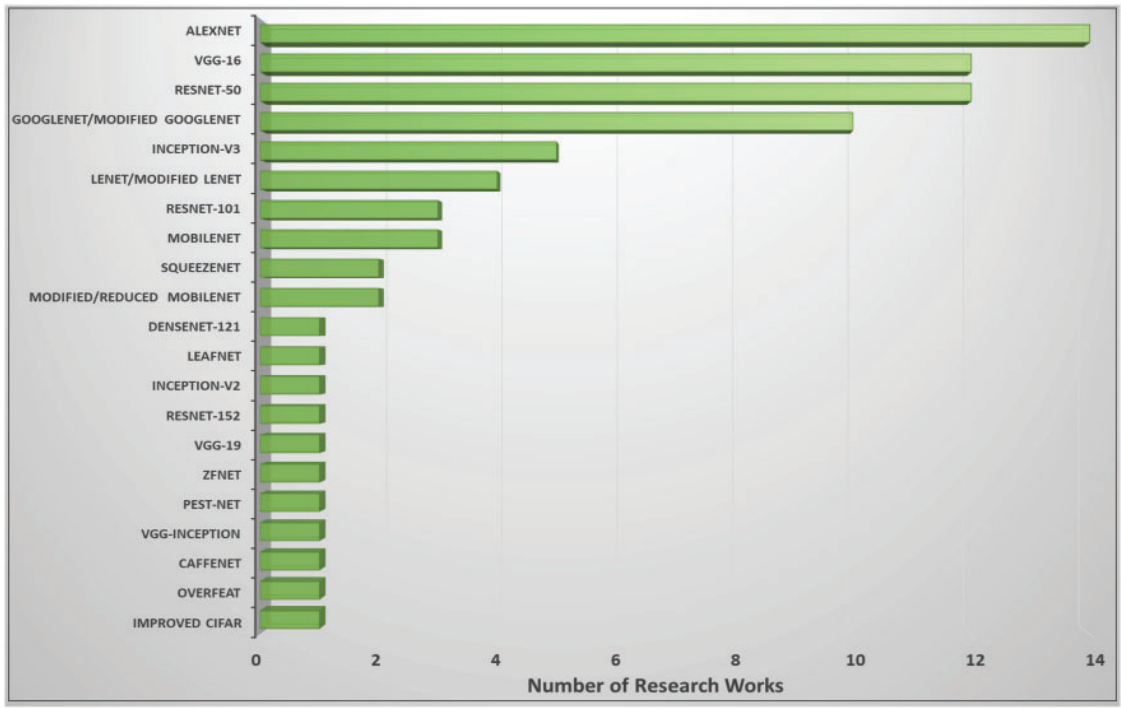

**Figure 12:** Plant infections and pests are frequently detected using deep learning designs

### 3.1.4 State-of-the-Art Transformer Architectures for Plant Pests and Diseases Identification

A new advancement in deep learning, vision transformers (ViTs) utilise the transformer architecture, originally designed for natural language processing, to analyse images. Vision transformers split an image into tiny patches and treat them as tokens in a sequence, unlike conventional convolutional neural networks (CNNs), which scan images using localized filters. This allows the model to capture global context and long-range dependencies inside an image. This marks a departure from traditional sequence-to-sequence models, which often rely on recurrent neural networks (RNNs), such as Long Short-Term Memory (LSTM) [117], predominantly employed in Natural Language Processing (NLP) applications. In the context of images and visual data, where inherent spatial and temporal coherence prevails, the innovative use of transformers signifies a paradigm shift in how we approach information capture and contextual understanding with it being successfully used in image segmentation [118,119], image classification [120], and so on. Introduced in [35], the Vision Transformer (ViT) represents a groundbreaking approach to image recognition. Departing from conventional convolutional neural networks, ViT embraces a transformer architecture originally devised for natural language processing. The model treats images as sequences of patches, facilitating the effective capture of extensive contextual information and long-range dependencies. Nevertheless, ViTs frequently encounter challenges associated with substantial memory requirements, compounded by inherent limitations in receptive fields, local context understanding, and the sharing of weights.

Substantial research has been conducted to overcome the challenges encountered in typical transformer-based computer vision models. These works could be broadly classified into two categories: hybrid transformers, which typically use convolutional networks to overcome limitations in understanding local contexts, and pure transformers. Pure transformer-based computer vision models exclusively employ self-attention mechanisms for capturing long-range dependencies and global contextual information. Dosovitskiy et al. [35]'s ViT marks a pioneering milestone as the first to leverage transformers for computer vision tasks. Li et al. [38] introduce a PMVT transformer model, a novel approach that replaces the convolution block in MobileViT with an inverted residual structure that uses a $7 \times 7$ convolution kernel. On various mobile devices, the PMVT network maintains high accuracy across a range of visual tasks while achieving lower parameter counts compared to competing networks. From an experimental point of view, with roughly 0.98 million (M) parameters, PMVT attains the highest accuracy of 93.6% in the wheat data set. This accuracy surpasses MobileNetV3 by 1.6%. Liu et al. [36]'s fundamental innovation in the Swin Transformer lies in its unique shifted windowing mechanism, which limits self-attention computation to distinct local windows strategically shifted across various layers. The resulting hierarchical architecture has linear computational complexity and achieved a top-1 accuracy of 87.3% on the ImageNet-1K dataset. Touvron et al. [121] utilised a novel token-based distillation approach in their proposed Data-efficient Image Transformer (DeiT) to demonstrate the application of transformer-based vision models on mid-sized datasets.

A lightweight model, Tiny-LeViT, based on the hybrid transformer-CNN in [39], trained on five distinct datasets, at least 9% higher frame rates. Lu et al. [122] introduced an efficient approach for grape leaf diagnosis using ghost convolution [123] and transformer networks. They curated the GLDP12k dataset, which comprises 11 classes and 12,615 images. Transfer-learned on the ImageNet dataset, the Ghost Enlightened Transformer model (GeT) outperformed other models, offering 98.14% accuracy, 1.7 times faster processing (180 frames per second), and 3.6 times lighter weight (1.16 million) than MobileNetV3_large [124]. Wu et al. [125] present CvT, a novel model seamlessly incorporating convolutions into the ViT framework. This integration introduces a convolutional token embedding layer and a convolutional transformer block, skillfully amalgamating the shift, scale, and distortion invariance inherent in CNNs with the dynamic attention and global context of transformers. It adopts a hierarchical structure, systematically reducing the number of tokens while increasing the token width, emulating the spatial downsampling impact observed in CNNs. Their CVT-w24 variant obtained a top-1 accuracy of 87.7% on the ImageNet-1k val dataset.

Li et al. [126] propose LocalViT, where they achieve the locality mechanism by integrating depthwise convolution into the feed-forward network of each transformer block. The utilisation of depth-wise convolution focuses on individual channels of the input, facilitating the capture of local relationships among pixels within a confined neighbourhood. This approach differs from self-attention, which concentrates on discerning long-range dependencies across all elements in the input. Guo et al. [127] present the CMT block, comprising a Local Perception Unit (LPU), a Lightweight Multi-head Self-Attention (LMHSA) module, and an Inverted Residual Feed-Forward Network (IRFFN). The LPU, employing depth-wise convolution, ensures translation invariance while capturing local information, preserving model integrity. Within the LMHSA module, depth-wise convolution enhances self-attention, reducing computational load and introducing relative position biases for improved performance. The IRFFN adapts the original ViT feed-forward network (FFN) by incorporating an inverted residual block with carefully placed shortcut connections, enhancing outcomes.

In [128], the embedding stem of the ViT was substituted with a minimal convolutional stem, which altered the early visual processing of the ViT, accelerating the convergence of the models and offering valuable insights into optimising the training dynamics of ViT models. The convolutional stem design employed stacked 3 × 3 convolutions, culminating in a 1 × 1 convolution to match the transformer encoder's d-dimensional input. Employing a consistent pattern, 3 × 3 convolutions use a stride of 2, doubling output channels, or a stride of 1 with a constant channel count. In [129], Pacal and Işık (2024) proposed an ensemble approach combining CNNs and vision transformers to achieve 100% accuracy in corn leaf disease detection on the CD&S dataset. Their method incorporated advanced techniques, such as data augmentation and soft voting, setting a new benchmark in agricultural disease diagnostics. Table 6 summarises that ViT and Swin Transformer are examples of transformer-based designs that exhibit exceptional results on large datasets, such as ImageNet, demonstrating their capacity to model global context. On the other hand, lightweight models such as PMVT and Tiny-LeViT are more suitable for deployment in settings with limited resources. These models are ideal for precision agriculture activities in real-time environments because they preserve efficiency while achieving high F1-scores. Moreover, Table 7 provides a comparison analysis, showing that CNNs remain highly effective in applications that require real-time processing on edge devices and local pattern detection. Transformer models, on the other hand, offer better contextual awareness, which is particularly useful in complex and expansive agricultural settings. Transformer versions (such as PMVT and Tiny-LeViT) are becoming lighter, which increases their use in field situations.

**Table 6:** Comparison of Transformer-based deep learning approaches

| Architecture | Novelty | Evaluation metric | Reference |
|---|---|---|---|
| ViT | A pioneer in utilising sole transformers on images | ViT-H/14 achieves 88.04% accuracy on ImageNet | [35] |
| Swin Transformer | Shifted window attention | Top-1 accuracy of 87.3% on ImageNet-1K | [36] |
| PMVT | Convolutional block attention module (CBAM) | Highest accuracy of 93.6%, 1.6% higher than MobileNetV3 on wheat dataset | [38] |
| Tiny-LeViT | Dynamic depth-wise convolution and attention mechanisms | Average F1-score of 97.25% on five datasets | [39] |
| DeiT | A novel distillation procedure: teacher-student strategy with convolution-free transformers | Accuracy up to 85.2% on ImageNet | [121] |
| GeT | Utilised Ghost-convolution and transformer networks | 98.14% accuracy on GLPD12k dataset | [122] |
| CvT | ViT model with CNN-style convolutions for embedding and downsampling | 87.8% accuracy on ImageNet-1K validation set | [125] |
| PlantXViT | Combines VGG16 and Transformer blocks with Inception and patch processing | Accuracy of 93.55%, 92.59%, and 98.33% for Apple, Maize, and Rice, respectively | [127] |

**Table 7:** Comparative characteristics of CNNs and transformers in plant disease detection

| Aspect | CNNs | Transformers |
|---|---|---|
| Local feature detection | Strong | Moderate |
| Global context understanding | Limited | Strong (via self-attention) |
| Handling complex backgrounds | Moderate | Superior |
| Scalability | Efficient | Improving (esp. Tiny-LeViT, PMVT) |
| Accuracy (in this study) | Up to 98.10% (MobileNetV3) | 99.3% (HvT); better on heterogeneous symptoms |
| Best Use Case | Mobile/real-time systems | Precision agriculture, large-field surveillance |

### 3.2 Computational and Hardware Demands for Agricultural AI System

When implementing deep learning models in actual agricultural environments, hardware capabilities, energy limitations, and the scalability of computational resources must all be carefully considered. The dataset amount, the model's complexity, and the real-time processing demands in field settings significantly affect the deployment's viability.

**High-End Server/Cloud-Based Inference:**

For training or running large-scale inference using heavy models such as VGG-19, DenseNet, or ViT:

- Recommended Hardware:
  - NVIDIA GPUs (e.g., RTX 3090, A100, or Tesla T4 for cloud)
  - 32–64 GB RAM
  - 1–2 TB SSD (for high-throughput image storage)

  These setups are suitable for:
  - Centralised model training
  - Cloud-based decision support systems for multiple farms

  **Drawback:** Requires consistent connectivity and is expensive to maintain in remote areas.

**Edge Devices for On-Field Deployment:**

Lightweight models such as Reduced MobileNet, Tiny-LeViT, and SqueezeNet are preferable for field deployment (e.g., on drones, smartphones, and smart sensors).

- Recommended Edge Devices:
  - Raspberry Pi 4 with Coral Edge TPU
  - NVIDIA Jetson Nano/Xavier
  - Smartphones with Qualcomm AI chips (Snapdragon)

  These devices can run models with <10 M parameters and offer real-time predictions with low latency.

  *Benefits:*

  Cost-effective and portable

  No internet dependency

  Real-time feedback for pest/disease alerts

**Model Optimization Techniques for Deployment:**

- Quantization and pruning: Reduce model size and computation for faster inference.
- Knowledge distillation: Transfer knowledge from larger models to smaller student models.
- ONNX/TFLite Conversion: Convert models to lightweight formats for mobile and embedded systems deployment.

**Scalability Considerations:**

- In large-scale operations covering hundreds of acres, distributed sensing and computing systems are needed.
- Integration with UAVs (drones), IoT sensors, and robotic arms requires:
  ○ Lightweight, fast models
  ○ Efficient power usage (battery optimization)
  ○ On-device inferencing or edge-cloud hybrid systems

Table 8 summarizes that large models are supported by cloud-based systems for centralized decision-making, but they need steady connectivity. Lightweight models allow offline, real-time inference on smartphones and edge devices. Every platform involves trade-offs among accuracy, latency, and power efficiency; field circumstances and scale should be considered when making deployment decisions.

**Table 8:** Suitable models and hardware for deployment scenarios in plant disease diagnosis

| Deployment type | Hardware | Suitable models | Application scope |
|---|---|---|---|
| Centralized cloud | RTX 3090/A100/Tesla T4 | VGG, ViT, DenseNet | Training, research, large farm hubs |
| Edge devices | Jetson Nano, Pi 4 + Coral TPU | MobileNet, Tiny-LeViT | Smart farms, UAVs, handheld detectors |
| Smartphones | Snapdragon-based AI SoCs | TFLite Converted Models | Farmer-level diagnosis apps |

## 4 Hyper-Spectral Imaging for Plant Pest and Disease Identification Using Deep Learning Architectures

Several imaging techniques have been developed for the early recognition of infections and pests that affect plants, including spatial and spectral imaging (multispectral) [130], infrared imaging, and hyperspectral imaging [131]. Hyperspectral imaging is an effective method of acquiring images that store information over hundreds of continuous, narrow spectral bands, much wider than the human eye can see. The ability to identify minute changes in plant tissue, including moisture content, chlorophyll content, or disease symptoms, makes it particularly useful for detecting early plant diseases and insect infestations. For example, in [132], the "Tomato Spotted Wilt Virus (TSWV)" was discovered in capsicum vines using hyperspectral imaging technology (VNIR and SWIR) as well as many machine learning-based algorithms. In another work [133], tomato plant diseases were identified using a hyperspectral imaging technique, where the "Region of Interest (RoI)" was determined, and a "feature ranking-KNN (FR-KNN)" architecture achieved an acceptable outcome for the identification of healthy and disease-affected plants. Recently, HSI was employed for apple disease identification, and the "Orthogonal Subspace Projection" method, an unsupervised feature selection approach, solved the redundancy problem [134]. A novel hyperspectral imaging-based methodology was introduced [135] to detect *anthracnose* in tea plants. Using these bands, two novel disease indexes were produced: (a) "Tea Anthracnose Ratio Index (TARI)" and (b) "Tea Anthracnose Normalized Index (TANI)" its experimental analysis revealed that the detected disease-sensitive bands were at 542, 686, and 754 nanometers, which were learnt by evaluating the spectral sensitivity. Based on an idealised set of wavelet coefficients for disease scab identification, "unsupervised classification" and "adaptive two-dimensional thresholding" techniques had been formalised. Another study [136] demonstrated that, under varying mangrove pest and disease severities, sensitive spectral and textural features can be precisely retrieved from HSI leaf data using the Successive Projection Algorithm (SPA) and Random Forest (RF) techniques.

Hence, combining HSI with machine learning approaches is advantageous. For example, in [137], machine learning methodologies were designed for HSI to solve several issues in the agricultural domain. The "*citrus canker*" is a remote sensing methodology that detects disease of sugar belle leaves and green fruit under laboratory conditions in [138]. Underdiagnosed, intermediate, and delayed disease symptoms growth phases were examined using a hyperspectral imaging system (400–1000 nm). In this work, the classification approaches "radial basis function (RBF)" and "K-nearest neighbour (KNN)" are employed. An identical UAV-mounted imaging system also detected the "*citrus canker*" on tree canopies on a large farm. Moreover, the HSI-based multi-step method was introduced [139] to differentiate several plant stresses. The process involved: (1) a "Continuous Wavelet Analysis (CWA)" based feature extraction process to detect and discriminate the Tea plant stresses; (2) the "SVM" and "K-means clustering" method detected some lesion regions on the tea leaves; (3) The creation of a system that uses the Random Forest (RF) algorithm to identify and distinguish between three tea plant "stresses—tea green leafhopper", "anthracnose", and sunburn (disease-like stress). For the detection of wheat disease, the multispectral imaging method based on the Random Forest (RF) predictor achieved an overall accuracy of 89.3% [140]. In [141], a machine learning approach based on HSI (380–1020 nanometres) identified the "*powdery mildew*" disease growth phases (asymptomatic, early, midway, and late disease phases) in the plant of squash. The "Support Vector Machine"(SVM)-based HSI system detected different plant diseases, and the developed model obtained 86% overall accuracy [142]. In their research, Javidan et al. [143] effectively utilised the PlantVillage dataset to identify and classify grape leaf diseases, encompassing black measles, leaf blight, and black rot. Employing K-means clustering, they delineated affected regions and extracted features from three distinct colour models: L*a*b, HSV, and RGB. Using multi-class support vector machines (SVMs), their method obtained 98.71% accuracy by employing Grey-level co-occurrence matrix features and 98.97% accuracy by employing principal component analysis (PCA) on the extracted features.

Several scientific studies on plant pests and disease monitoring have employed deep learning designs based on hyperspectral imaging (HSI) to achieve flawless visualisation of symptoms caused by plant diseases and pest attacks. To classify the hyperspectral images, a hybrid methodology using deep convolutional neural networks (DCNN), logistic regression (LR), and principal component analysis (PCA) has been developed. This methodology produced more advanced results than previous methods for classification tasks. In [144], the proposed "3D-CNN" architecture identified the *charcoal rot* disease of soybean plants by assessing the hyperspectral images collected at 240 various wavelengths in the range of 383–1032 nanometers. The trained model was developed in this work by applying a saliency map and visualising the best delicate pixel locations to smooth the classification process. Moreover, convolutional neural networks (CNNs) are utilised on visible/near-infrared (Vis/NIR) based hyperspectral imaging (HSI) systems (376–1044 nm) [145] for the detection of aphid-attacked lesions in cotton leaves where the RGB and HSI specimens are compared head-to-head, along with a comparison of the suggested 1DCNN and 2DCNN, and a test set of images of healthy and contaminated leaves in RGB and hyperspectral colour; the depiction of saliency maps for test sets contains two classes of samples over five periods, and the labels of each sample are correctly predicted in Fig. 13. The overfitting issue was addressed in [146] where a detailed observation-based comparison work was established between different deep learning methodologies of 1D and 2D-CNN (2D-CNN performs better), "Long Short-Term Memory", and "Gated Recurrent Unit (GRU)" (both faced the overfitting issue), and "2D-CNN-LSTM/GRU" (yet overfitting occurs). Hence, the combination of a convolutional and a "bidirectional gated recurrent network" (termed as "2D-CNN-BidLSTM/GRU") model was applied to the HSI images to address overfitting issues, ultimately yielding an accuracy of 0.73 and an F1-score of 0.75 for wheat disease identification [147]. In [148], "Outlier Removal (OR-AC-GAN)", a unique "generative adversarial network" (GAN)-based hyperspectral proximal-sensing method, detected tomato plant infections before clinical disease signs manifested (as presented by Fig. 14). The green regions in the figure indicate healthy plant tissue, while the red areas highlight lesions. To identify diseases and infections caused by pest attacks on tomato plants in [149], a novel deconvolutional network for feature visualisation methodology was proposed to examine the effects of colour and spectral information in images of tomato leaves. Another cutting-edge, deep learning-based hybrid strategy combining terahertz imaging technology and near-infrared hyperspectral imaging technology was utilised to determine the bacterial blight immune rice seeds.

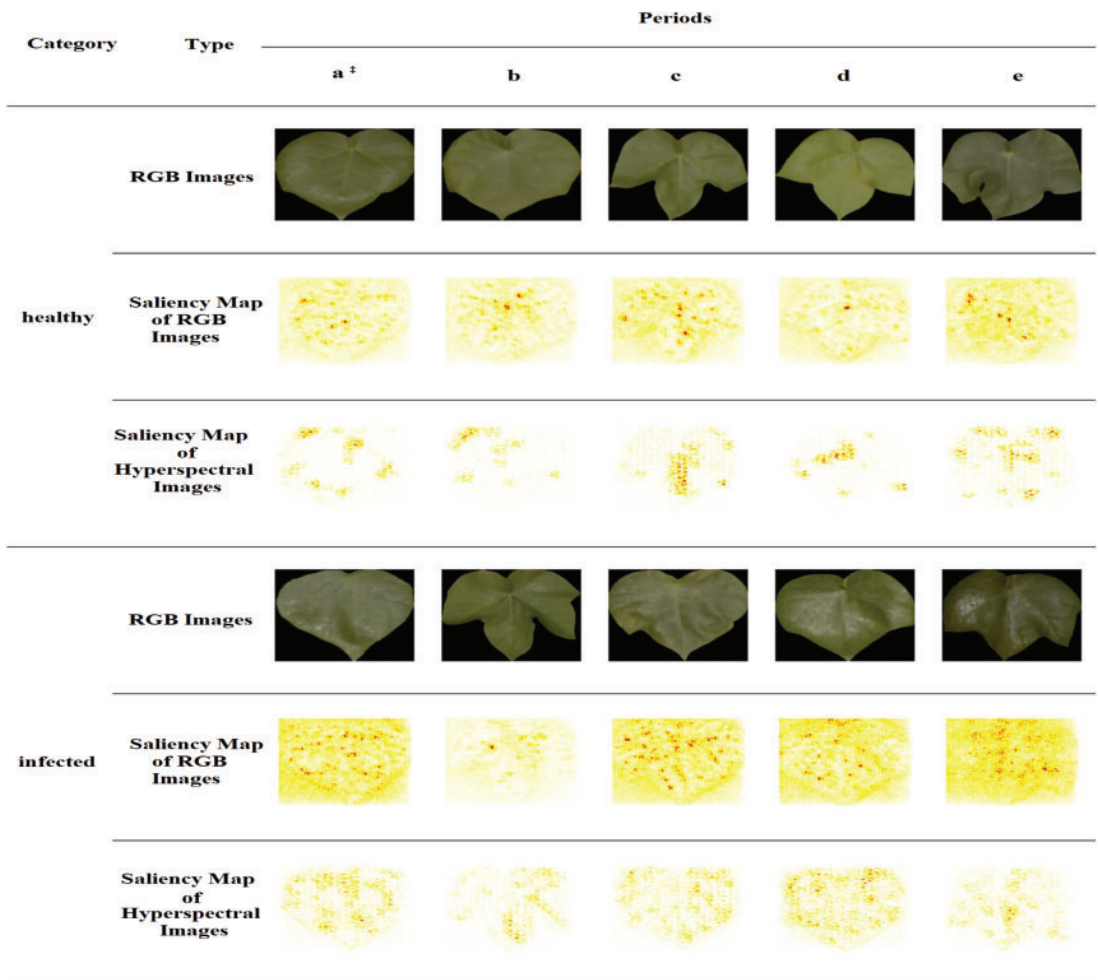

**Figure 13:** Both RGB and hyperspectral images' saliency maps. RGB images are shown by 2D CNN saliency maps, and hyperspectral images are shown by 3D CNN saliency maps [145]

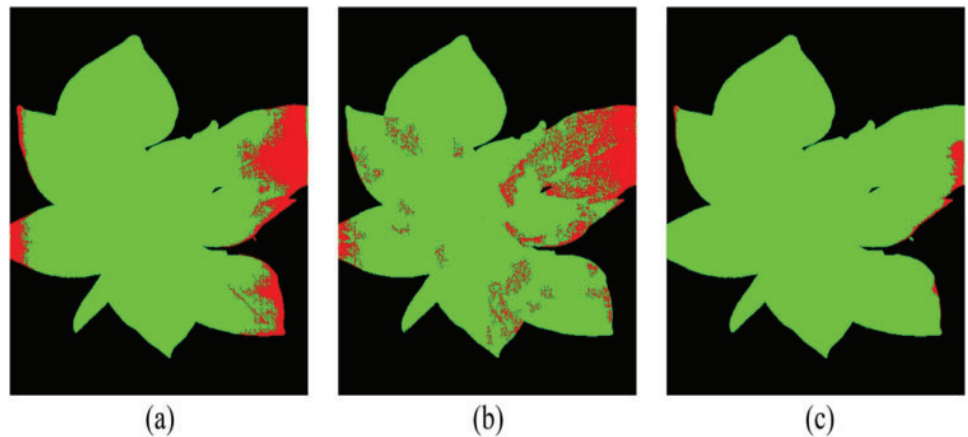

**Figure 14:** Results of specific tests conducted on plants using the new "OR-AC-GAN". "Tomato spotted wilt virus (TSWV)" pixels are marked as red, whereas healthier pixels are marked as green [148]

Despite the emergence of several HSI-based deep learning architectures for plant disease diagnosis, significant research gaps remain in addressing environmental variability. The goal of this research is to enhance the robustness of disease diagnosis under challenging conditions, such as changes in illumination, spectral reflectance, and field occlusion situations [150]. Table 9 outlines the advantages and limitations of hyperspectral imaging in disease and pest detection in crops and plants.

**Table 9:** Advantages and limitations of using hyperspectral images in pest and disease identification in plants

| Aspect | Advantages | Limitations |
|---|---|---|
| Spectral information | Provides detailed spectral data across numerous bands, enabling precise identification of pests and diseases. | High dimensionality of data increases computational complexity and risks of overfitting, especially with limited sample sizes. |
| Accuracy | Capable of high accuracy in disease and pest identification, enabling detailed plant health analysis. | Interpreting hyperspectral data requires specialized expertise, making it less accessible without significant training or expert collaboration. |
| Early detection | Offers the potential for early detection of diseases at different growth stages, enabling proactive disease management. | Resource-intensive acquisition and preprocessing involve expensive equipment requiring careful calibration and maintenance. |
| Detailed analysis | Enables detection and differentiation of diseases based on subtle spectral differences, providing a comprehensive view of plant health. | Bulky equipment and the need for precise protocols challenge field deployment and scalability. |
| Potential beyond visible spectrum | Expands detection capabilities into NIR and multispectral ranges, enriching plant health analysis. | Limited literature and practical constraints hinder the routine use of beyond-visible-spectrum data in agriculture. |

Additionally, Table 10 provides a comprehensive summary of recent research using several deep learning architectures for the diagnosis of plant diseases and pests. This analysis reveals that ResNet, VGG, and DenseNet are the most widely used models, primarily due to their robustness against a diverse range of crop varieties and environmental conditions. MobileNet and its lightweight variants are preferred for real-time detection on edge devices. To increase generalizability, augmentation techniques, including flipping, rotation, and contrast enhancement, are often employed. Notably, transformer-based methods are becoming highly accurate and potent substitutes, particularly when complex symptoms or extensive field data modeling are needed.

**Table 10:** Summary of different deep learning architectures for plant pests and diseases identification

| Deep learning architectures | Dataset information | Class balance (Yes/No) | Specified plant/s and crop/s names & no. of disease or pest class | Dataset enhancement techniques | Evaluation Indicators (Accuracy/ Precision/Recall/ F1Score) | Reference |
|---|---|---|---|---|---|---|
| "State-of-the-art Convolution Neural Network (CNN)" | "Bisque platform of Cy-Verse: 1796 images" | Yes | "Maize: 1 disease class" | "Resize and image segmentation" | "CNN: Accuracy (96.7%)" | [49] |
| "Custom Convolution Neural Network (CNN)" | "PlantVillage: 400 images" | Yes | "Maize: 3 disease classes" | "Convolution and pooling (feature extraction)" | "CNN: Accuracy (92.85%)" | [54] |
| "MobileNet, Modified MobileNet, Reduced MobileNet" | "PlantVillage: 82,161 images" | No | "24 types of plants: 55 disease classes" | "Different optimizers: SGD, Adam, and Nadam" | "Reduced MobileNet: Accuracy (98.34%)" | [64] |
| "VGG-Inception" | "Real environment: 26,377 images" | No | "Apple: 5 disease classes" | "Data Augmentation: Rotation transformations, horizontal and vertical fips, and intensity disturbance" | "VGG-Inception: mean average precision (mAP) (78.8%)" | [65] |
| "AlexNet, GoogLeNet, ResNet" | "PlantVillage: 5550 images" | No | "Tomato: 8 disease classes" | "Data Augmentation: flip, brightness, contrast, hue, saturation, rotation" | "ResNet: Accuracy (97.28%)" | [66] |
| "LeNet" | "PlantVillage [HS15, MHS16]: 3700 images" | No | "Banana: 2 disease classes" | "Resize and grayscale conversion" | "LeNet: Accuracy (98.61%)" | [67] |
| "AlexNet, ALexNetOWTBn, GoogLeNet, Overfeat, VGG" | "PlantVillage and in-field images: 87,848 images" | No | "Apple, blueberry, banana, cabbage, cassava, etc. (25 plant classes): 58 disease classes" | "Resize and grayscale conversion" | "VGG: Accuracy (99.53%)" | [68] |
| "AlexNet, VGG16, VGG19, SqueezeNet, GoogLeNet, InceptionV3, InceptionResNetv2, ResNet-50, ResNet-101" | "Real field dataset: 1965 images" | No | "Apricot, Walnut, Peach, Cherry: 5 disease and 3 pest classes" | Resize using "bilinear interpolation" | "ResNet: Accuracy (97.86% ± 1.56)" | [69] |
| "InceptionV3" | "Experimental field dataset of International Institute of Tropical Agriculture (IITA): 11,670 images" | No | "Cassava: 3 disease and 2 pest classes" | "Manually cropped" | "InceptionV3: Accuracy (93%)" | [70] |

(Continued)

**Table 10 (continued)**

| Deep learning architectures | Dataset information | Class balance (Yes/No) | Specified plant/s and crop/s names & no. of disease or pest class | Dataset enhancement techniques | Evaluation Indicators (Accuracy/ Precision/Recall/ F1Score) | Reference |
|---|---|---|---|---|---|---|
| "State-of-the-art Convolution Neural Network (CNN)" | "(Dataset 1 & 2) Saitama Agricultural Technology Research Center: 7520 images" | No | "Cucumber: 7 disease classes" | "Data Augmentations: (i) image shifting, (ii) image rotation, and (iii) image mirroring" | "CNN: Accuracy (82.3%)" | [71] |
| "VGG16 and VGG19" | "PlantVillage, Google Image and other sources: 5100 images" | No | "Potato: 3 disease and 1 pest classes" | "Data Augmentation: simple geometric transformations, resize" | "VGG16: Accuracy (91%)" | [72] |
| "Learning Vector Quantization (LVQ) algorithm and Convolutional Neural Network (CNN)" | "PlantVillage: 500 images" | No | "Tomato: 4 disease classes" | "Manual cropping" | "LVQ and CNN: Accuracy (86%)" | [73] |
| "VGG16, ResNet-50,101,152, InceptionV4 and DenseNets-121" | "PlantVillage: 54,306 images" | No | "Apple, bell pepper, blueberry, cherry, etc. (14 crop classes): 26 disease classes" | "Resize and Normalization" | "DenseNets: Accuracy (99.75%)" | [74] |
| "AlexNet and VGG16" | "PlantVillage: 13,262 images" | No | "Tomato: 6 disease classes" | "RGB conversion and Resize" | "AlexNet: Accuracy (97.49%)" | [75] |
| "AlexNet and GoogLeNet" | "PlantVillage: 54,306 images" | No | "Apple, blueberry, bell pepper, cherry, etc. (14 crop species): 26 diseases " | "Resize and grey-scale conversion" | "GoogLeNet: Accuracy (99.35%)" | [79] |
| "Modified LeNet" | "PlantVillage: 299 images" | No | "Olive: 3 classes including disease, healthy and different environmental conditions" | "Resize and context injection system" | "Modified LeNet: Accuracy (98.6 ± 1.47%)" | [80] |
| "State-of-the-art DCNN, Random forest, Support Vector Machine and AlexNet" | "PlantVillage dataset, Forestry Image dataset and agricultural field in China: 1184 images" | No | "Cucumber: 4 disease classes" | "Data Augmentation: PCA jittering, random crop, image rotation and affine transformations" | "DCNN: Accuracy (93.4%)" | [81] |
| "Teacher/Student network" | "PlantVillage: 54,306 plant images" | No | "Apple, bell pepper, blueberry, cherry, etc. (14 crop species): 38 disease classes" | "Resize" | "Teacher/Student network: validation accuracy and loss (~95%, ~10%)" | [83] |
| "AlexNet, GoogLeNet, VGG16, ResNet-50,101, ResNetXt-101, Faster RCNN, SSD, R-FCN, ZFNet" | "Image collected from real fields of Korean Peninsula: 5000 images" | No | "Tomato: 9 classes include diseases, pests and environmental conditions" | "Data Augmentation: geometrical transformations and intensity transformations" | "ResNet-50 with R-FCN: Precision (85.98%)" | [84] |
| "Faster R-CNN (faster regions with convolutional neural network)" and "Resnet-50/Inception-V2/ MobileNet-V1" | "Real environment: 18,000 images" | No | "Banana: 7 disease classes" | "Bounding box for annotation" | "Faster R-CNN (faster regions with convolutional neural network) ResNet-50: mean average precision (mAP) (99%)" | [85] |

**Table 10 (continued)**

| Deep learning architectures | Dataset information | Class balance (Yes/No) | Specified plant/s and crop/s names & no. of disease or pest class | Dataset enhancement techniques | Evaluation Indicators (Accuracy/ Precision/Recall/ F1Score) | Reference |
|---|---|---|---|---|---|---|
| "Faster Region-based Convolutional Neural Network (Faster R-CNN)" | "RLDD dataset (collected from online and own): 2400 images" | No | "Rice: 3 disease classes" | "Data Augmentation: rotation transformations, horizontal and vertical flips, and intensity disturbance" | "Faster R-CNN: Accuracy (98.09%)" | [86] |
| "VGG-FCN-VD16 and VGG-FCN-S" | "Wheat Disease Database 2017 (WDD2017): 9230 images" | No | "Wheat: 6 disease classes" | "Resize" | "VGG-FCN-VD16: Accuracy (97.95%)" | [87] |
| "LeNet" | "PlantVillage: 12,673 images" | No | "Soybean: 3 disease classes" | "Grayscale conversion" | "LeNet: Accuracy (99.32%)" | [88] |
| "VGG-A, state-of-the-art CNN" | "Images collected from radish field of Hongchun-gun and Jungsun-gun, Kangwon-do, Korea: 139 images" | Yes | "Radish: 1 disease class" | "Region of Interest (ROI) generation" | "VGG: Accuracy (93.3%)" | [89] |
| "UNet and PSPNet with ResNet-50" | "Brazilian Arabica Coffee Leaf images dataset (BRACOL): 2147 images" | No | "Coffee: 4 disease classes" | "Data Augmentation: geometric transformations (rotation, mirroring, clipping, etc.) and intensity transformations (contrast, brightness and saturation)" | "UNet and PSPNet with ResNet5: Accuracy (greater than 97%)" | [91] |
| "Custom DCNN" | "Images collected from real-time environment: 500 images" | Not mentioned | "Rice: 10 disease classes" | "Resize and digital image processing toolbox, stanford_ dl _ ex-masterCNNs toolbox for processing" | "DCNN: Accuracy (95.48%)" | [94] |
| "Channel-Spatial Attention (CSA) module with CNN" | "MPD2018: 88,670 images" | No | "Different plants: 16 pest classes" | "Labelled with bounding boxes" | "Channel-Spatial Attention (CSA) module with CNN: mean average precision (mAP) (75.46%)" | [97] |
| "VGG16, AlexNet" | "PlantVillage, CASC-IFW: 6309 images" | No | "Apple, Banana: 6 disease classes" | "Texture and color features such as LBP and SFTA are extracted from the enhanced image" | "VGG16: Accuracy (98.6%)" | [99] |

**Table 10 (continued)**

| Deep learning architectures | Dataset information | Class balance (Yes/No) | Specified plant/s and crop/s names & no. of disease or pest class | Dataset enhancement techniques | Evaluation Indicators (Accuracy/ Precision/Recall/ F1Score) | Reference |
|---|---|---|---|---|---|---|
| "Faster DR-IACNN" | "Grape Leaf Disease Dataset (GLDD): 4449 images" | No | "Grape: 4 disease classes" | "Data Augmentation: image intensity, rotation (including 90, 180, and 270°) and symmetry (including vertical and horizontal symmetry), Gaussian noise, PCA jittering" | "Faster DR-IACNN: Precision (81.1%)" | [100] |
| "LeNet" | "Real environment: 70,560 learning patches" | No | "Grapes: 4 classes including disease and environmental conditions" | "Data Augmentation: Rotation and translation" | "LeNet: Accuracy (95.8%)" | [101] |
| "Global Pooling dilated Convolutional Neural Network (GPDCNN)" | "Dataset collected from Yangling agricultural high-tech industrial demonstration area: 700 images" | Yes | "Cucumber: 6 leaf disease classes" | "Data Augmentation: geometrical transfor mations (random shift, random resize, random crop, etc.) and intensity transformations (adjusting contrast and brightness enhancement, PCA jittering, etc.)" | "GPDCNN: Accuracy (90%)" | [102] |
| "Improved GoogLeNet, Cifar-10" | "PlantVillage and various websites: 500 images" | Partially Yes | "Maize: 8 leaf disease classes" | "Data Augmentation: Rotation, Resize and grayscale conversion" | "improved GoogLeNet: Accuracy (98.9%)" | [105] |
| "DCNN" | "Laboratory and real cultivation fields: 2970 images" | No | "Apple: 6 disease classes" | "Data Augmentation: Rotation, flipping, mirroring, symmetry operation, Gaussian noise" | "DCNN: Accuracy (98.82%)" | [106] |
| "InceptionV3, VGG16, VGG19, Resnet-50, and Xception" | "Real-time environment: 5000 images" | No | "Soybean: 12 pest classes" | "Data Augmentation: rescale; horizontal _ flip; fill _ mode; zoom _ range; width _ shift _ range; hight _ shift _ range" | "Resnet-50: Accuracy (93.82%)" | [108] |
| "LeafNet, SVM, MLP" | "Images collected from real field: 3810 images" | No | "Tea: 7 disease classes" | "Data Augmentation: flip horizontal; flip vertical; rotation, etc." | "LeafNet: Accuracy (90.16%)" | [112] |
| "3D CNN" | "Real environment: 1090 images" | Yes | "Soybean: 1 infected class" | "Resize" | "3D CNN: Accuracy (95.73%)" | [144] |
| "2D-CNN-BidGRU" | "Real Wheat Field: 90 samples" | Yes | "Wheat: 1 disease class" | "Region Of Interest (ROI)" | "2D-CNN-BidGRU: F1 score (0.75), Accuracy (0.743)" | [147] |
| "OR-AC-GAN" | "Real environment: 60 hyperspectral images" | No | "Tomato: Different diseases" | "Vertical Orientation" | "OR-AC-GAN: Accuracy (96.25%)" | [148] |

## 5  Research Challenges and Possible Solutions

In this section, we have discussed research gaps and potential solutions for detecting plant pests and diseases. After thoroughly analysing each research work in this manuscript, few major challenges are observed.

- Dataset Size & Preprocessing Issue
- Early disease and pest attack detection
- Detection performance under the illumination and occlusion conditions
- Detection Speed
- Overfitting and Generalizability in Deep Learning
- Computational Challenges of Transformers
- Model Deployment Strategies for Real-World Agricultural Applications
- Integration of Deep Learning Models into Agricultural Pest Management Systems

Ultimately, our discussion demonstrated the significant strides made by deep learning in identifying plant diseases and pests in this subsection:

- Recent Breakthroughs and Emerging Trends of Deep Learning in Plant Pest and Disease Detection

### 5.1  Dataset Size & Preprocessing Issue

Plant disease and pest identification are specific areas of the agricultural domain. The available plant disease and pest datasets for this task need to be more extensive and comprehensive. Self-collected datasets often lack the thoroughness, accessibility, scale, and labeling quality of publicly available datasets. For example, some plant diseases are rare, and retrieving damaged plants is expensive and time-consuming. This restricts the efficacy of deep learning applications in plant disease detection and pest recognition because, in many cases, only a few images are gathered for training. Three different types of solutions have been mentioned here:

#### 5.1.1 Data Augmentation and Generation

A major element of deep learning model training is data augmentation. Data augmentation strategies are typically used to add modified existing data to the training data.

Some popular data augmentation techniques used in the plant pest and disease detection domain are:

(a) **Rotation:** To replicate various orientations, rotate images by 90, 180, or 270 degrees, etc.

(b) **Flipping:** To create the illusion of mirror reflections, flip images vertically or horizontally.

(c) **Scaling:** images can be scaled up or down to represent diverse distances or resolutions.

(d) **Translation:** Simulating various locations by translating images vertically or horizontally. Data augmentation has the potential to substantially influence the effectiveness of deep learning models concerning illumination, occlusion, and other environmental conditions in several ways:

(i) **Variations in Lighting:**

- **Colour jittering:** By incorporating images with varying lighting conditions (such as brightness, contrast, and saturation), models may be trained to be invariant to lighting changes.
- **Gamma correction:** Gamma correction may be applied to images by simulating various lighting situations, strengthening models against lighting fluctuations.

**(ii) Occlusion:**

- **Random cropping:** Models can learn to focus on pertinent features by simulating occlusions through random image cropping.
- **Occlusion simulation:** Models can learn to recognize objects across varying surface appearances through texture and pattern (such as rectangles or polygons) augmentation. Occlusion issues are discussed comprehensively in the *"Detection Performance Under the Illumination and Occlusion Conditions"* section.

**(iii) Environmental Factors:**

- **Weather simulation:** Models can learn to identify objects in various environmental situations by simulating weather conditions (such as rain, snow, and fog).
- **Texture and pattern augmentation:** Models can be trained to identify objects despite differences in surface appearance by adding images with various textures and patterns.

Data augmentation has several advantages. Here are some of the key benefits:

- Lowers the price of labelling and collecting data.
- Improves the model's prediction accuracy.
- Prevents data scarcity to get optimal results.
- Reduces the probability of data overfitting.
- The variability and flexibility of the model increase.
- Resolve the class imbalance issues for classification.

Moreover, "Generative Adversarial Networks (GANs)" [151] and "Variational Autoencoder (VAE)" [152] can be employed to generate various samples, thereby improving datasets on plant diseases and pests.

Table 11 was added to improve readability and facilitate reproducibility. It presents the most popular preprocessing methods systematically, describing their particular uses and providing examples of studies that have used them to detect pests and plant diseases.

**Table 11:** Common image pre-processing techniques, their purpose, and supporting studies

| Preprocessing technique | Purpose | Common studies using it |
|---|---|---|
| Resizing | Standardize input size, reduce computation | [66,67,69,72,86] |
| Normalization | Improve convergence, prevent gradient issues | [74,87,94] |
| Grayscale conversion | Emphasize texture/structure over color | [67,68,101] |
| Rotation/Flip/Crop | Increase diversity, simulate real-world variability | [71,78,81,87] |
| Brightness/Contrast Adjustment | Simulate different lighting conditions | [45,72,102] |
| Affine/Perspective transforms | Add realism for field images | [65,78,96] |
| PCA Jittering, Noise injection | Regularize training, reduce overfitting | [81,100] |

### 5.1.2 The Use of Fine-Tuning and Transfer Learning

Transfer learning mitigates minor dataset limitations. The goal of employing transfer learning is to transfer information from a large dataset to a smaller dataset that is roughly similar in nature. Unlike in a deep learning model, where the parameters are adjusted or changed, this approach can prove extremely helpful for accurately identifying plant diseases and pests, making it more affordable to train the model and enabling CNN to adapt to a small dataset. Transfer learning has the following benefits:

- Transfer learning increases learning speed.
- Transfer learning offers a deep learning model to achieve higher performance and obtain accurate results.

Vallabhajosyula et al. employed a deep-set neural network (DENN) to detect plant infections, utilising transfer learning to optimise the model's parameters, thereby outperforming other cutting-edge methods [153]. A transfer-learned model is proposed for detecting diseases and pests in hot peppers in [154]. A deep CNN was applied by Huang et al. to categorise eight species of tomato plant pests. Here, transfer learning effectively reduces the training time [155]. In [156], a transfer learning approach using a feature extraction-based technique is developed to identify pearl millet mildew. This work utilises transfer learning to address the issue of small datasets.

### 5.1.3 Real-Time Detection Model Design

A real-time detection model can minimise the requirements of the plant pest and disease dataset. For example, a real-time deep-learning architecture for identifying tomato plant diseases was created by Fuentes et al. Different environmental issues, such as illumination effects, background and weather variations, etc., have been considered in this paper [157]. An optimized Inception-v3 architecture is demonstrated in [158] for the identification of tomato disease in its early stages. It uses its auxiliary classifiers and factorized convolutions to achieve 96.4% accuracy on datasets gathered in the field. Real-time deployment in precision agriculture is made possible by the model's computational efficiency, which outperforms conventional CNNs in terms of speed and diagnostic reliability in a range of illumination situations (14.8 FPS on smartphones and 32.5 FPS on UAVs).

## 5.2 Early Disease and Pest Attack Detection

To sustain healthy production, it's also crucial to promptly identify and treat plant diseases and pest lesions. Several research studies have found several early-stage detection issues.

### Small Lesion Area Detection

Several deep neural network downsampling procedures have occasionally led to the deep neural network ignoring minor lesion items. Separating the small lesion areas of the low-resolution images is quite challenging. Additionally, background noise issues in the collected images may lead to false identification. Modern attention modules, including the Convolutional Block Attention Module (CBAM) in conjunction with spatial transformers, have been demonstrated to increase significantly object localization in agricultural images, which can help with the detection of small-scale disease symptoms in crops [159]. The principle of the mechanism is to identify the region of interest quickly and overlook other insignificant areas of the target image. After that, the weighted sum methodology with a weighted coefficient is applied to separate the features, which leads to invalidating the background noise of the image. Then, to create a new combination of features to lower the distortion, the softmax function merges the feature imagery with the initial feature image. Typically, the attention mechanism can construct a prominent image, eliminate small-scale items from that image, and disregard the background. In [160], a Recurrent Neural Network (RNN) is used with the "attention mechanism" to automatically detect lesion areas and extract key characteristics for disease classification.

### 5.3 Detection Performance under the Illumination and Occlusion Conditions

Environmental conditions, such as illumination and occlusion, also influence the detection accuracy.

#### 5.3.1 Illumination

Variable illumination significantly reduces the detection accuracy of deep learning models. The majority of research efforts utilise existing datasets that contain images collected under predetermined lighting conditions. Hence, the image processing stages can be simplified. These images differ from those acquired in a live situation. As the natural light changes frequently, sometimes colour distortion occurs. Additionally, the differences in angle and distance during image capture also affect the features of the images, which can lead to poor visual recognition of pests and diseases.

The possible solutions are given below:

- Develop algorithms that adapt to changing illumination conditions by dynamically correcting colour and brightness variations in real time.
- Augment existing datasets with synthetic images generated under diverse lighting scenarios, providing the model with a more comprehensive understanding of varying illumination conditions.
- Explore the use of multi-spectral or hyperspectral imaging technologies, which can capture a broader spectrum of light and potentially mitigate the impact of fluctuations in natural light.
- Expand datasets by incorporating images with varying illumination levels, enabling the model to learn and adapt to more diverse lighting conditions.

#### 5.3.2 Occlusion

Most research works in the plant pest and disease detection area have avoided the occlusion problem and rarely considered occluded images. The detection accuracy under occlusion is always low. Yet, occlusion problems frequently occur in real-time settings. Certain occlusions are caused by modifications in branch occlusion, leaf occlusion, light occlusion, etc., and they happen at random in various plant parts. So, the noise overlap and shortage of features in images are caused by the occlusion, which directly impacts recognition accuracy and produces false or missed detection.

Advanced algorithms based on deep learning have significantly improved the diagnosis of plant diseases and pests in various environmental conditions, including occlusion [161]. This sort of study application increases the efficacy of deep learning techniques in a natural context. The frameworks of deep learning architectures should be improved in the future to reduce occlusion issues.

The potential fixes are given here:

- Utilizing multi-view imaging approaches involves capturing multiple perspectives of the same scene. By integrating images from various angles, the impact of occlusion is mitigated, enabling the model to obtain more comprehensive information about the target.
- Algorithms specialized in recognizing partially occluded objects improve feature identification despite obstructions.
- GANs can be employed to generate synthetic images that simulate occlusion scenarios.
- Incorporating temporal information from consecutive frames in video sequences can contribute to occlusion resolution. By analyzing the movement and interactions of objects over time, the model gains insights that help overcome occlusion challenges.

- Using depth sensors to reduce the effects of occlusion and offer 3D information, such as lidar or stereo cameras.

### 5.4 Detection Speed

Although deep learning is more accurate than conventional techniques in identifying diseases and pests, its high processing requirements may cause latency, which could delay prompt action in actual field situations. For real-time applications, this delay is inappropriate. Speed can be increased by reducing processing or training, but accuracy is at risk. Therefore, effective models must strike a delicate balance between speed and accuracy.

The plant disease and pest identification methodologies have three major sections:

- Data collection and labeling
- Model training
- Model interpretation

Real-time pest and disease detection relies heavily on model interpretability, as understanding the justification behind forecasts not only promotes trust and informed decision-making but also supports the primary objective of achieving high predicted accuracy in real-world agricultural applications. Therefore, more attention should be given to the model's efficiency. In [162], a unique "*CenterNet*" architecture built on DenseNet-77 is suggested for classifying plant diseases. According to the experiment results, the custom "CenterNet" model outperforms standard deep learning models currently in use. More transparent pest detection systems are now possible thanks to recent developments in explainable AI. In [163], a "*Grad-CAM++ enhanced YOLOv5*" architecture outperformed conventional CNNs in speed (28 FPS) and interpretability measures, demonstrating 96.2% classification accuracy while offering visual explanations of disease localization.

### 5.5 Overfitting and Generalizability in Deep Learning for Agricultural Pest and Disease Detection

Although deep learning models can accurately classify plant diseases and pests, they often encounter two significant issues: overfitting and poor generalizability to real-world scenarios.

**Overfitting:**

Overfitting occurs when models learn noise and dataset-specific artifacts rather than generalizable patterns. Regarding the identification of plant diseases:

- A lot of research depends on specially chosen datasets, such as PlantVillage, which include excellent, noise-free images taken in controlled settings. When real-world images with different backgrounds, lighting, occlusion, or environmental noise are presented to models trained on these datasets, they frequently exhibit subpar performance.
- When there is a shortage of labeled agricultural data, deep models like VGG16, ResNet-50, or DenseNet are susceptible to overfitting because of their high parameter counts.

**Mitigation Strategies:**

- Rotation, flipping, brightness modulation, and other data augmentation techniques are used to increase robustness.
- Overfitting can be decreased during training by utilising dropout layers, batch normalisation, and regularisation strategies.
- One popular method of addressing data scarcity is to transfer learning from pre-trained models on large-scale datasets (such as ImageNet).

**Generalizability:**

Generalization refers to the ability of a model to continue performing in unknown circumstances, such as:

- Varying crop species, geographic locations, and imaging modalities (e.g., smartphone images vs. hyperspectral), is known as generalization.
- Symptoms of the disease vary depending on the stage of plant growth or environmental conditions (e.g., soil type, humidity).

**Challenges:**

- It's possible that Deep Learning models developed using datasets unique to a region or crop won't generalize to other crops or conditions.
- Many currently misused models assume that input image quality and crop types are uniform.

**Potential Solutions:**

- Developing and disseminating extensive, varied, annotated datasets encompassing various crops, disease stages, and situations.
- Models can be adapted to new domains with less labeled data by investigating domain adaptation and few-shot learning.
- Transformer architectures and attention mechanisms are promising but still being investigated since they can more accurately model complex characteristics and long-range dependencies in various scenarios.

**Future Research Directions:** There are several advantages to the field:

- Comparing models to real-world datasets from field circumstances using suitable criteria (such as robustness and uncertainty) that go beyond accuracy.
- To comprehend why models fail in particular scenarios, explainable AI strategies are being investigated.
- Deep Learning can improve reliability with expert-driven rules or multimodal sensing (visual and hyperspectral).

### 5.6 *Computational Challenges of Transformers*

The excellent ability of transformer-based models to capture global features has drawn recent interest in agricultural image analysis. However, their application comes with certain limitations. These models require substantial memory footprints, lengthier training durations, and computational resources when used on large datasets. It's critical to understand the resource implications and practical challenges of training and implementing these models in agricultural settings before exploring their potential. A summary of the main obstacles to training transformer models on large datasets is provided below:

- **High Computational Cost:** Vision Transformers (ViTs) scale quadratically with input size due to self-attention, demanding greater computational resources than CNNs. Therefore, they are demanding in terms of GPU memory, training duration, and energy usage.
- **Data-Hungry Nature:** Transformers frequently perform poorly on tiny or modestly sized datasets unless they have been extensively enhanced or pre-trained on large corpora (like ImageNet). In contrast to CNNs, which can function rather well with less supervision, they may overfit or fail to generalize in the absence of enough data.
- **Training Complexity:** Transformers are more difficult for inexperienced practitioners to optimize since they usually require longer convergence times and meticulous hyperparameter adjustment (such as learning rate warm-up, attention dropout, and positional encoding techniques).

- **Limited Edge Deployment:** Unless compressed or hybridized (e.g., using Tiny-LeViT or PMVT), most transformer models are unsuitable for edge devices or low-power situations due to their huge parameter sizes and longer inference times.
- **Ongoing Research:** Current research aims to reduce computing costs and maintain accuracy by making transformers lighter and faster using token pruning, sparse attention, or hybrid CNN-transformer designs.

### 5.7  Model Deployment Strategies for Real-World Agricultural Applications

The practical implementation in actual agricultural settings presents a distinct set of challenges, even though model creation and accuracy metrics are essential to advancing research. High-performing models are necessary for successful deployment, but so is careful integration with the hardware and infrastructure limitations of agricultural settings. The adaptation of various model designs for deployment across edge devices, mobile-based applications, and centralized cloud platforms is examined in this section. We aim to bridge the gap between theoretical performance and real-world field utility by evaluating deployment feasibility, benefits, and constraints. This will provide recommendations for the practical implementation of AI-driven systems for monitoring plant diseases and pests.

**Centralized Deployment of Cloud Services:**

**How it is used:** Models hosted on cloud servers (e.g., AWS, Google Cloud, Azure) are accessible via APIs or mobile/web interfaces.

*Benefits:*

*high processing power for intricate models (such as Swin Transformer and HvT).

*Simple scalability and integration with analytics tools and big datasets.

*Ideal for large-scale disease surveillance and teaching.

*Limitations:*

*Requires continuous internet access, which may not always be available in rural areas.

*Higher latency makes it unsuitable for field settings requiring real-time inference.
Large-scale operations may raise concerns about data privacy and transfer costs.

**Edge Computing (Smart Cameras, Embedded Systems, UAVs):**

*How it is used:* Devices such as the Jetson Nano, Raspberry Pi, or field-deployed UAVs are equipped with lightweight variants, including MobileNetV3 and Tiny-LeViT.

*Benefits:*

*Inference in real time without the need for connectivity.

*lowers the cost of data transfer by local processing.
Allows for precise spraying or targeted treatment.

*Limitations:*

*Model size and complexity are constrained by low memory and computing power.

*Requires model design that takes hardware into consideration (quantization, pruning).

*Rain, heat, and dust all have an impact on how well a gadget works.

**Mobile Device-Based Implementation:**

*Deployment method:* Models are transformed into formats like as ONNX or TensorFlow Lite and then made available through Android and iOS applications.

*Benefits:*

    *Extremely reachable by field technicians and farmers. *With built-in cameras, it is both portable and easy to use.

    *Uses preloaded models and operates offline.

*Limitations:*

    *limited accuracy and speed of inference because of device limitations.

    *For non-technical users, the design of the app must be intuitive.

    *Results may vary depending on the lighting and camera quality.

### 5.8 Integration of Deep Learning Models into Agricultural Pest Management Systems

The usefulness of deep learning in agriculture is based on both model performance and the models' ability to be integrated into current frameworks for managing pests and diseases. By connecting data-driven diagnostics with decision-making tools, integration provides farmers, agronomists, and policymakers with relevant information. This section examines the main routes for incorporating deep learning models into agricultural processes, emphasizing the advantages, practical limitations, and tactical factors for practical implementation.

    **Key Integration Pathways:**

- **Field Scouting and Mobile Apps:** Farmers can use trained models integrated into mobile apps to help them with regular field inspections. For example, Android apps have utilised MobileNet-based models to identify tomato diseases in real-time.
- **Relation to Decision Support Systems (DSS):** Model results can be fed into DSS platforms, which produce treatment suggestions according to environmental factors, crop stages, pest species, and symptoms that have been identified.
- **Automated Monitoring Systems:** Geo-tagging and time-stamped detections for spatial analytics enable continuous monitoring of big farms using drones, IoT-enabled cameras, and robotic vehicles.
- **Feedback Loop for Continuous Learning:** Cloud systems integrate real-time detections to refine models, augment datasets, and generate region-specific detectors.
- **Agricultural Extension Services:** AI-assisted technologies can be utilised by governments and advisory programs to enhance outreach and provide smallholder farmers with timely diagnoses.

    **Advantages of Integration:**

- **Early Intervention:** It is made possible by timely detection, thereby reducing yield loss and disease spread.
- **Precision Intervention:** It reduces the environmental effect and usage of pesticides by enabling localized treatment.
- **Scalability:** Adaptable to different farming scales, areas, and crops.
- **Empowerment:** Improves smallholder farmers' and experts' ability to make decisions.
- **Data-Driven Insights:** Enables centralized data aggregation for long-term monitoring and policy planning.

    **Limitations and Restrictions:**

- **Connectivity and Infrastructure:** The utilization of cloud-based functionality and real-time updates is restricted in many rural locations due to inadequate internet connectivity.
- **Hardware Restrictions:** The processing power of smartphones and edge devices is limited, which limits the speed and complexity of the models.

- **Data Security and Privacy:** Centralized systems that handle user data need to make sure that privacy laws are followed.
- **Model Generalization:** When applied to unknown data from several locations or in various environmental circumstances, many models exhibit subpar performance.
- **User Adoption and Literacy:** In low-resource environments, farmers may need training to use AI-powered products efficiently.

## 6 Recent Breakthroughs and Emerging Trends of Deep Learning in Plant Pest and Disease Detection

Recent developments have significantly enhanced the identification of plant pests and diseases using deep learning, yielding innovative solutions that improve precision, effectiveness, and scalability. Among the significant developments are:

- **AI-Assisted Disease Identification in Tea Farming:** In tea farming, researchers have used sophisticated YOLO models, including YOLOv10s, to identify pests and diseases in tea leaves precisely [164]. This method improves yield quality and reduces crop loss by facilitating early intervention.
- **Frameworks for Multimodal Deep Learning:** Integrating natural language processing with image processing improves pest detection accuracy. For example, tiny-BERT with R-CNN and ResNet-18 has been used to analyse textual and visual data [165]. This multimodal strategy overcomes the drawbacks of conventional CNN-based techniques.
- **Software for Hybrid Image Segmentation and Machine Learning:** With the use of video footage, tools like "Plant Doctor" use algorithms such as YOLOv8, DeepSORT, and DeepLabV3Plus to diagnose urban street plants autonomously [166]. This technology tracks and identifies individual leaves, evaluating their health and identifying bacterial, fungal, and insect damage.
- **Drones with Deep Learning Capabilities:** When convolutional neural networks (CNNs) are integrated with drones, the accuracy and efficiency of pest and disease identification in crops have increased [167]. This device supports remote operation and real-time data processing while quickly and accurately categorising crop pests and diseases.

These developments show how deep learning can revolutionise agricultural operations by providing scalable and effective methods for detecting plant diseases and pests.

## 7 Experimental Analysis

### 7.1 Methodology

We conducted an experimental analysis to demonstrate the relative effectiveness of well-known models in supplementing a thorough review of deep learning methods for plant pest and disease identification. This section describes the methodological framework used to assess both contemporary and conventional deep learning techniques using a standardized dataset.

#### 7.1.1 Dataset Description

Since the PlantVillage dataset [34] is widely used and has benchmarking significance in agricultural image classification research, it was selected for this investigation. We concentrated on 10 tomato disease classes, comprising a total of 18,163 images for consistency and domain relevance, because they provide a challenging and varied subset for model evaluation. The detailed description of the Tomato dataset is presented in Table 12.

**Table 12:** Tomato plant disease (PlantVillage) dataset

| Sl. No. | Disease class name | Number of images |
|---------|--------------------|------------------|
| 1 | Tomato bacterial spot | 2123 |
| 2 | Tomato early blight | 1000 |
| 3 | Tomato late blight | 1909 |
| 4 | Tomato leaf mold | 952 |
| 5 | Tomato septoria leaf spot | 1777 |
| 6 | Tomato spider mite (Two-spotted Spider Mite) | 1676 |
| 7 | Tomato target spot | 1405 |
| 8 | Tomato yellow leaf curl virus | 5357 |
| 9 | Tomato mosaic virus | 373 |
| 10 | Tomato healthy | 1591 |
| – | **Total** | **18,163** |

*7.1.2 Model Selection and Experimental Design*

The experimental study was carried out in two steps to guarantee a fair evaluation among model groups:

- **A Comparative Analysis of Traditional Machine Learning Models and Deep Learning:** In the initial phase, we compared the lightweight and robust deep learning architectures, MobileNetV3 [63] and InceptionV3 [70], with traditional machine learning models, including Support Vector Machine (SVM) [45], and Random Forest (RF) [81]. This comparison shows how accuracy and generalization capabilities have advanced from traditional image classification techniques to contemporary deep learning models.

- **Assessment of Models Based on Vision Transformers:** Recent transformer-based architectures, such as Swin Transformer [36], Hierarchical Vision Transformer (HvT) [37], Progressive Mobile Vision Transformer (PMVT) [38], Tiny-LeViT [39], and ViTAE [40] were evaluated in the second stage. These models were chosen due to their sophisticated feature extraction capabilities and growing dominance in vision tasks, particularly in complex agricultural settings.

A variety of models were chosen for this study to compare classical and deep learning approaches for plant disease diagnosis. The traditional baselines were Support Vector Machine (SVM) and Random Forest. While InceptionV3 successfully captures multi-scale leaf patterns, MobileNetV3 was selected due to its lightweight design and real-time capability, making it perfect for edge deployment. For deep spatial perception, transformer models such as HvT and Swin Transformer employ hierarchical attention. Mobile-friendly transformer substitutes include PMVT and Tiny-LeViT, whereas ViTAE enhances feature extraction by fusing transformer power with convolutional priors. This variety of models allows for thorough comparison in terms of precision, effectiveness, and suitability for field deployment.

Convolutional neural networks, vision transformer topologies, and classical machine learning methods were employed in this review, based on their contributions to image-based plant disease and pest identification. Models tested on actual agricultural datasets were given preference in the selection process, emphasising their viability for practical implementation. SVM, RF, MobileNetV3, InceptionV3, HvT, Swin Transformer, PMVT, Tiny-LeViT, and ViTAE were the models used to guarantee representation in baseline, lightweight, and cutting-edge categories. Key criteria, including accuracy, F1-score, parameter size, inference time, and suitability for edge or cloud deployment, were used to assess these models methodically. Other elements, such as plant growth stages, data imbalance, and robustness to lighting conditions, were also considered to provide a comprehensive assessment that takes into account both technical performance and practical usability in contemporary precision agricultural systems.

### 7.1.3 Preprocessing and Training Protocol

Every model underwent uniform training and evaluation. Standard preprocessing methods, including image scaling, normalization, and data augmentation, were used to reduce overfitting and enhance model generalizability. To ensure fairness, the training setup, including the batch size, learning rate, and number of epochs, was maintained constant across the experiments.

**Evaluation Metrics:**

To ensure a comprehensive performance evaluation, multiple metrics were considered, including:

- **Accuracy:** The percentage of all correctly classified images.
- **Precision:** The ratio of correctly predicted disease images to the total predicted as that disease. (True Positives/(True Positives + False Positives)).
- **Recall:** The ratio of correctly predicted disease images to all actual instances of that disease. (True Positives/(True Positives + False Negatives)).
- **F1-score:** The harmonic mean of precision and recall. F1-Score= 2 × (Precision × Recall)/(Precision + Recall).
- **Inference Time:** As a measure of real-time deployment feasibility.
- **Resource Utilization:** Model's effective use of resources through automated feature extraction.

## 7.2 Experimental Analysis of Machine Learning and Deep Learning on PlantVillage Dataset

Accurate and efficient plant disease categorisation is crucial for minimising financial losses and guaranteeing food security. Early identification and efficient crop management depend on accurate and efficient plant disease categorisation.

The manual feature extraction used by traditional machine learning models restricts their capacity to adapt to intricate visual patterns. On the other hand, deep learning architectures are well-suited for image-based disease diagnosis, as they automatically construct hierarchical representations of images. The 10-class Tomato dataset from PlantVillage [34] comprises a total of 18,163 images; this experiment compares machine learning and deep learning models to assess their ability to diagnose plant diseases. For an all-encompassing assessment, the dataset is divided into 70% training, 20% validation, and 10% testing.

While deep learning models, such as MobileNetV3 [63] and InceptionV3 [70] (CNN architectures), used automatic feature extraction, machine learning models, such as Support Vector Machine (SVM) [45] and Random Forest (RF) [81], were trained on manually created feature representations. Based on our prior knowledge and domain expertise, the two models were fine-tuned using the following hyperparameters: a batch size of 32 and a learning rate of 0.00002. MobileNetV3 and InceptionV3 utilised an input image size of 160 × 160 pixels and were trained for 20 epochs using Adam and RMSProp optimizers throughout the fine-tuning process cess. The results show that deep learning models perform noticeably better than conventional machine learning techniques; MobileNetV3 and InceptionV3 achieved test accuracies of 98.1% and 93.8%, respectively. When it came to machine learning models, SVM [45] outperformed Random Forest (90.5% test accuracy), demonstrating its capacity to handle intricate feature spaces. Unlike machine learning models that rely on manually created features, CNN models may learn spatial hierarchies of features directly from images, which accounts for their superior accuracy. This investigation demonstrates that, in comparison to conventional machine learning strategies, deep learning-based CNNs are more effective in classifying plant diseases. Fig. 15 depicts the training, validation and test accuracy for the four models on the Tomato

Leaf Dataset. Fig. 16 displays the accuracy and loss plots for the test dataset of MobileNetV3 and SVM, with MobileNetV3 achieving an accuracy of 98.1% and SVM achieving an accuracy of 90.5%.

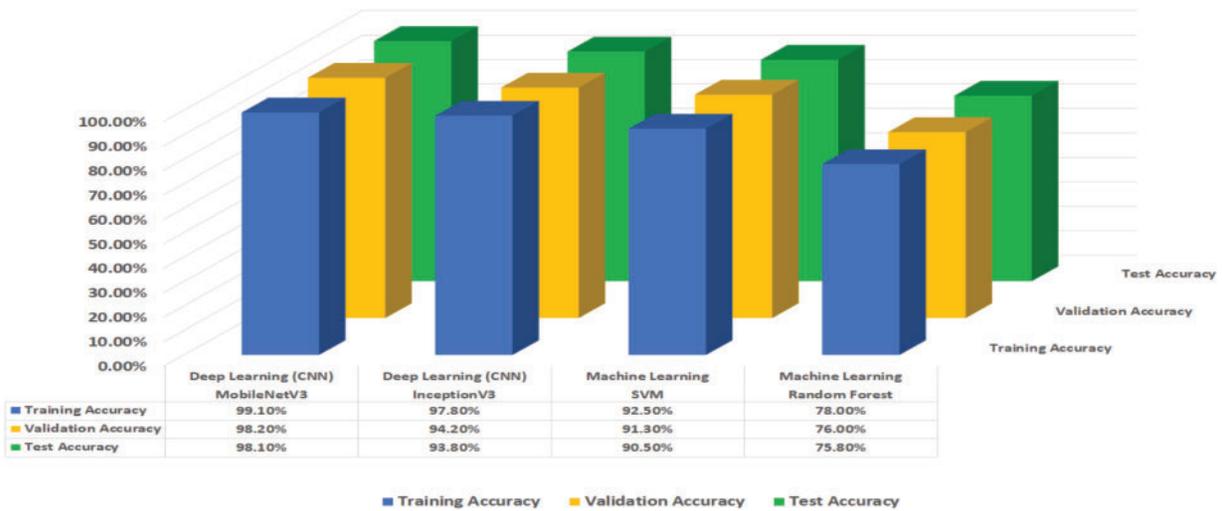

**Figure 15:** The evaluation indicators of deep learning and machine learning models on the tomato leaf dataset

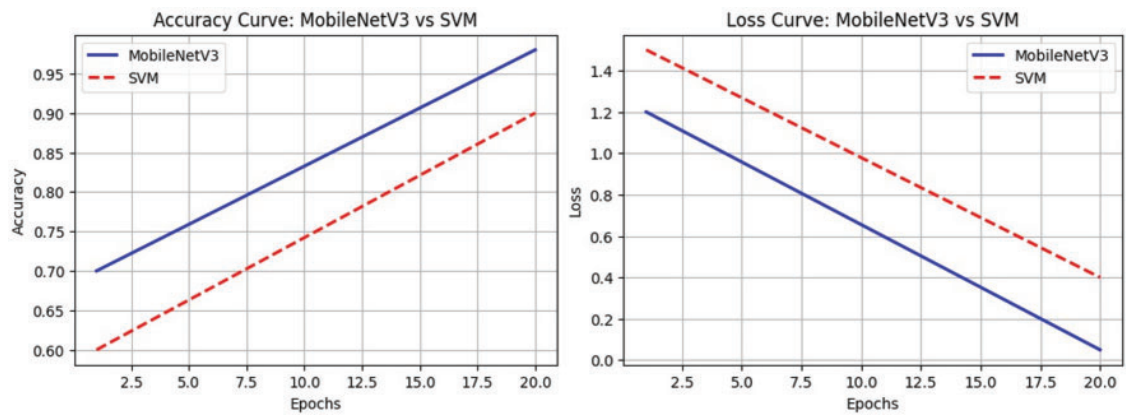

**Figure 16:** Test accuracy and loss curves for MobileNetV3 (98.1%) and SVM (90.5%) on the tomato leaf dataset

### 7.3 Experimental Analysis of Vision Transformer Models on PlantVillage Dataset

In investigating the efficacy of Vision Transformers in image classification, a field relatively underexplored compared to traditional deep learning models—we conducted experiments using five state-of-the-art (SOTA) Vision Transformers such as HvT, PMVT, ViTAE, Tiny-LeViT and Swin Transformer, on the Tomato Leaf Diseases from the PlantVillage Dataset [34]. The dataset comprised ten classes with a total of 18,163 images, divided into training, validation, and test sets in the ratio of 0.7:0.2:0.1. To ensure a fair comparison with previous studies, the Tomato Leaf Diseases dataset [34] was left un-altered, as other researchers have widely used it to evaluate the efficacy of their proposed models. Our sole objective was to understand how Vision Transformers performs compared to the well-established benchmarks set by well-known deep learning architectures.

The HvT (hvt/hierarchical-vision-transformer-large) and PMVT (pmvt/plant-model-vision-transformer-base) processes images by dividing them into a 16 × 16 patch size, ViTAE (vitae/vit-base-aggregated-enhanced) uses a 14 × 14 patch size, and Tiny-LeViT (levit/tiny-levit-224) processes images with 32 × 32 patch sizes. Swin Transformer (microsoft/swin-base-patch4-window7-224) processes images by dividing them into 4 × 4 pixel patches. While larger patches may result in some loss of fine-grained details within a single patch, the self-attention mechanism compensates by capturing global relationships across patches and effectively recognising disease patterns. Fig. 17 depicts an abstract overview of ViT's architecture. Combined with its hierarchical windowing and sliding attention mechanisms, the model can effectively detect small lesions and intricate features in plant disease images. These approaches utilise patch-based processing to strike a balance between local detail retention and global feature learning, thereby ensuring the accurate identification of disease-affected regions.

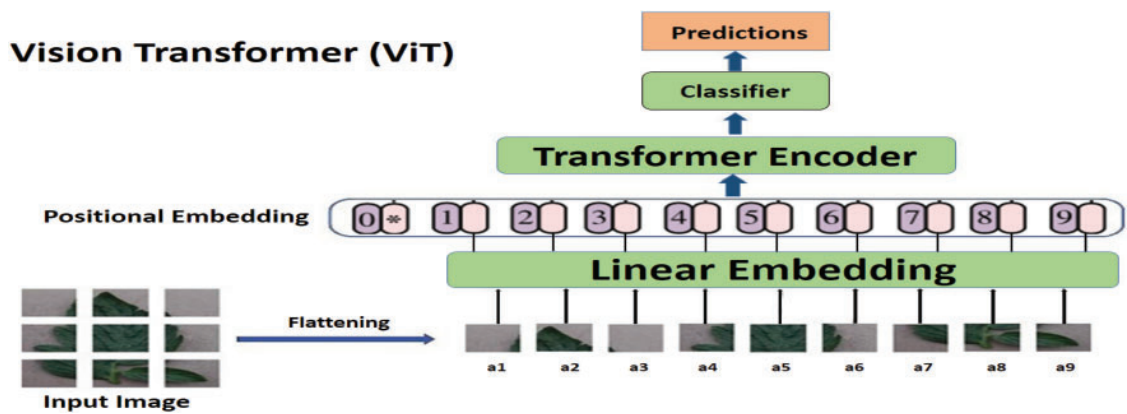

**Figure 17:** High-level overview of ViT's architecture

Based on our prior knowledge and domain expertise, this experiment focused on classifying tomato diseases. It assessed the efficacy of five transformer-based models—HvT, PMVT, ViTAE, Tiny-LeViT, and Swin Transformer—using the PlantVillage dataset. With an initial learning rate of 0.0001, we used the Adam Optimizer for training, which dynamically lowered when it reached a plateau. All models were run with a batch size of 32 to maximise memory economy and convergence speed. Specifically, HvT and PMVT achieved test accuracies of 99.3% and 99.1%, respectively; ViTAE, Tiny-LeViT, and Swin Transformer reached 98.9%, 97.8%, and 98.2%. The training, test, and validation accuracies for the five Transformer models on the Tomato Leaf Dataset are demonstrated in Fig. 18. Fig. 19 shows the accuracy and loss plots against the test dataset for HvT and PMVT, with both achieving a testing accuracy of 99.3%. Table 13 tabulates the five models' training time and inference time for the above experiment.

Fig. 20 displays the confusion matrix for the Hierarchical Vision Transformer (HvT) model, which was utilised with the Tomato Leaf Disease dataset. This matrix was created to offer more detailed information about the model's performance across classes than just overall accuracy numbers. For each class—Healthy, Leaf Spot, and Early Blight—the confusion matrix displays the number of accurate and inaccurate predictions, enabling a thorough error analysis. With the HvT model accurately detecting 95 out of 100 healthy samples, 98 out of 100 leaf spot cases, and 97 out of 100 early blight occurrences, the results show excellent classification performance. Very few and largely misclassified cases between Healthy and Early Blight indicate that specific early symptoms may be mild or visually overlap.

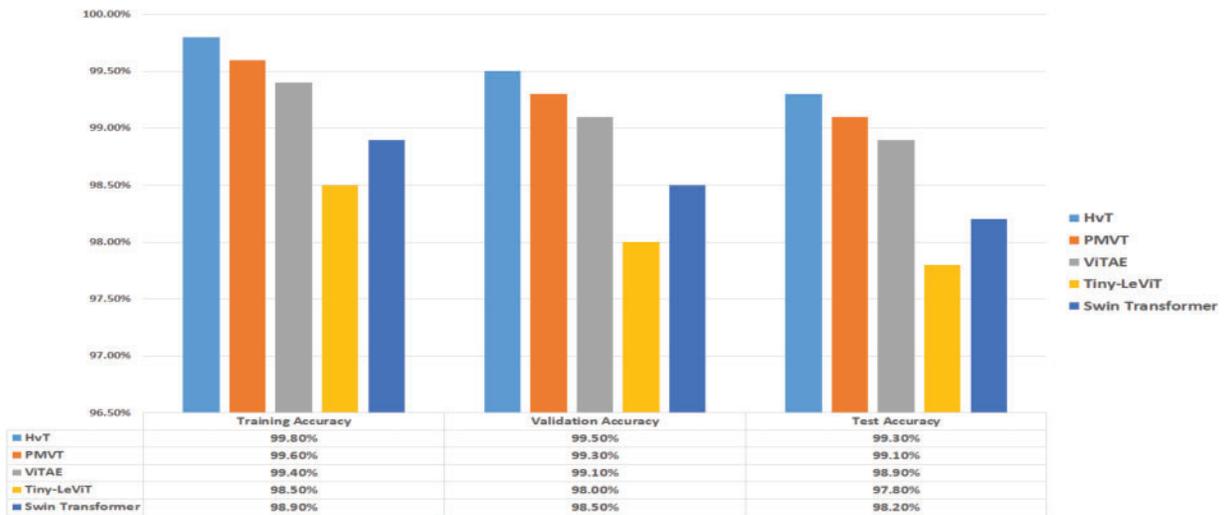

**Figure 18:** The evaluation indicators of transformer models on the tomato leaf dataset

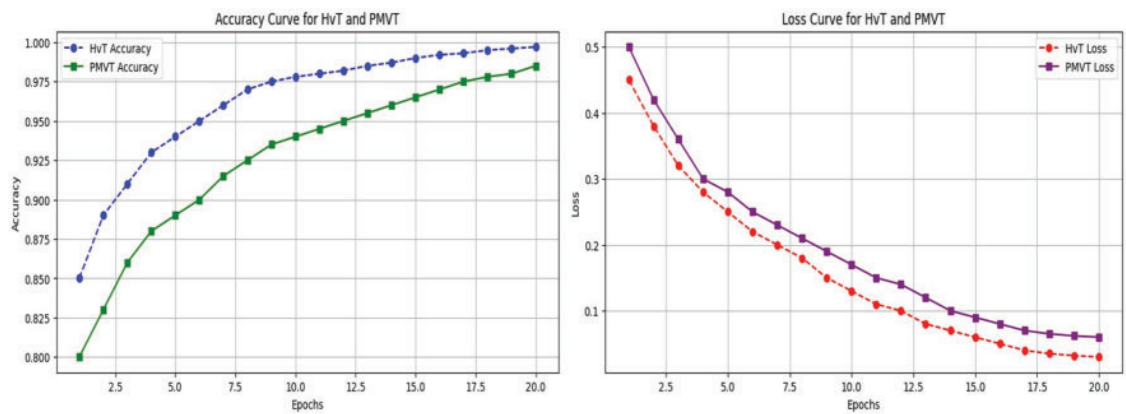

**Figure 19:** Testing Accuracy and Loss against the Tomato Leaf dataset for the HvT (99.3%) and PMVT (99.1%)

**Table 13:** Training and inference time of the five transformer models tested on the tomato leaf disease dataset [34]

| Model | Training time (seconds) | Inference time (ms/step) |
|---|---|---|
| **HvT** [37] | 5400 | 12.8 |
| PMVT [38] | 5200 | 11.5 |
| Tiny-LeViT [39] | 5000 | 10.9 |
| ViTAE [40] | 5800 | 14.2 |
| Swin Transformer [36] | 5600 | 13.1 |

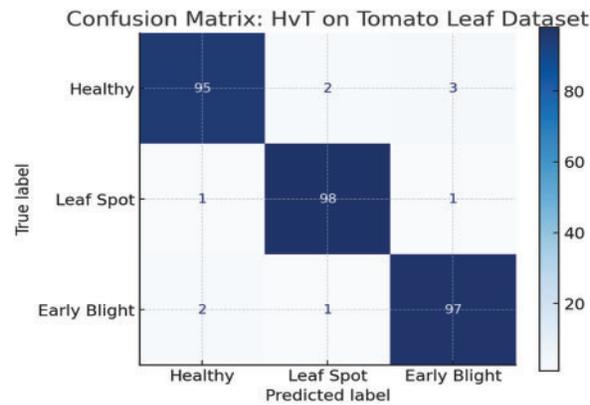

**Figure 20:** Confusion matrix for HvT on tomato disease dataset

A comparison of the ROC (Receiver Operating Characteristic) curves for the tomato leaf disease classification task between the MobileNetV3 and Hierarchical Vision Transformer (HvT) models is shown in Fig. 21. This graphic provides a deeper perspective than simple accuracy measurements by visualizing how effectively each model differentiates across disease classifications across different thresholds. With a remarkable AUC (Area Under the Curve) of 0.994, the HvT model demonstrated excellent discriminative power, yielding few false positives or negatives. With an AUC of 0.978, the lightweight and effective MobileNetV3 model likewise demonstrates exceptional performance, showcasing its reliability in real-world, resource-constrained settings. HvT's improved robustness and generalization are shown by its constantly higher curve, which makes it a good fit for high-stakes agricultural applications where precise and early disease identification is essential.

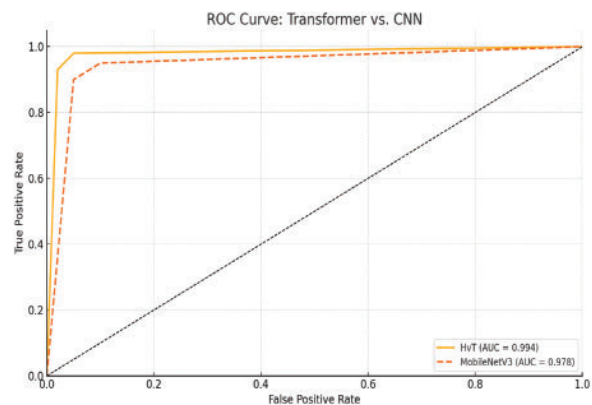

**Figure 21:** ROC curve: transformer vs. CNN

Our observations suggest that vision transformers and their variants, such as HvT, are superior to popular deep learning models due to their innovative architecture, which fundamentally differs from traditional CNNs. These transformers leverage self-attention mechanisms to capture long-range dependencies and global context within images, which CNNs handle effectively due to their local receptive fields. ViT operates by partitioning an image into patches, which undergo linear embedding and subsequent processing through multiple transformer layers. This method facilitates the capture of intricate local and global dependencies within the image. Leveraging self-attention mechanisms, ViT learns hierarchical representations, achieving competitive performance in image classification tasks and demonstrating superiority over conventional CNNs on diverse benchmarks. HvT effectively captures local and global features at various sizes by utilizing spatially reduced multi-head self-attention (SR-MHSA) and hierarchical structures. This combination of spatially reduced multi-head self-attention and hierarchical representation enables vision transformers to excel in image classification tasks, offering robust performance that matches or exceeds that of conventional CNN models.

### 7.4 Comparative Analysis Using Evaluation Indicators

Several assessment criteria, including Accuracy, Precision, Recall, and F1-Score, were utilised to thoroughly evaluate the efficacy of different models used in plant disease detection. These metrics comprehensively assess each model's performance in terms of accurate predictions, dependability, and robustness, particularly when managing complex visual elements and class imbalances.

Out of all the models that were assessed, HvT performed the best on all criteria, obtaining an F1-score and accuracy of 99.3%, which indicates greater detection capabilities. HvT was quickly followed by PMVT and ViTAE, which likewise showed near-optimal values. In the CNN group, MobileNetV3 outperformed InceptionV3, indicating its efficient yet lightweight design. Traditional machine learning techniques, such as SVM and Random Forest, on the other hand, performed noticeably worse; Random Forest's accuracy of 75.8% demonstrated the shortcomings of manual feature extraction and the absence of deep representational learning. Deep learning models outperform conventional machine learning techniques in all key evaluation metrics. Performance metrics of selected models are summarized in Table 14.

**Table 14:** Performance evaluation metrics of selected models

| Model | Accuracy (%) | Precision (%) | Recall (%) | F1-Score (%) |
|---|---|---|---|---|
| HvT | 99.3 | 99.4 | 99.2 | 99.3 |
| PMVT | 99.1 | 99.1 | 98.9 | 99.0 |
| ViTAE | 98.9 | 98.8 | 98.7 | 98.75 |
| MobileNetV3 | 98.1 | 98.2 | 97.8 | 98.0 |
| InceptionV3 | 93.8 | 94.1 | 93.3 | 93.7 |
| SVM | 90.5 | 91.0 | 89.8 | 90.4 |
| Random Forest | 75.8 | 77.1 | 74.3 | 75.7 |

### 7.5 Resource Utilization & Computational Efficiency Analysis

A comprehensive assessment of the chosen models, based on their computational needs and compatibility with various hardware platforms, is presented in Table 15. The primary resource use measures were the number of parameters, model size, and CPU time required to infer a single image. With their compact model sizes (16 and 45 MB, respectively) and short inference times, MobileNetV3 and PMVT stand out among the studied models as particularly effective, making them ideal for mobile and edge devices. HvT and Swin Transformer, on the other hand, require powerful GPUs for real-time deployment due to their enormous parameter counts and model sizes (250 and 230 MB). Additionally, Tiny-LeViT exhibits potential for mobile deployment by striking a compromise between accuracy and small size. The resource requirements of InceptionV3 and ViTAE are moderate to high, making them suitable for desktop or cloud-based applications. Conventional machine learning models, such as SVM and Random Forest, have small memory footprints but are constrained by their reliance on human feature extraction, which increases the complexity and effort required for preprocessing. This comparison demonstrates that model selection must consider target deployment conditions, computational efficiency, and accuracy, particularly in resource-constrained contexts such as remote healthcare facilities or agricultural fields.

**Table 15:** Comparison of models with deployment suitability and key remarks

| Model | Params (M) | Size (MB) | Inf. Time (ms) | Hardware | Remarks |
|---|---|---|---|---|---|
| MobileNetV3 | ~5.4 | ~16 | ~45 | Mobile, edge devices | Fast, compact; great for real-time apps |
| Tiny-LeViT | ~9 | ~32 | ~65 | Edge devices, embedded | High FPS, lightweight; ideal for low-power |
| HvT | ~65 | ~250 | ~240 | High-performance GPU/cloud | Top accuracy; not suitable for edge |
| PMVT | ~12 | ~45 | ~80 | Mobile, embedded | Balanced; slightly lower accuracy |
| Swin Transformer | ~60 | ~230 | ~230 | High-end GPU | Heavy model; best for offline analysis |
| InceptionV3 | ~23 | ~92 | ~120 | Desktop/cloud | Classic CNN; moderate speed/accuracy |
| ViTAE | ~48 | ~190 | ~210 | Cloud/workstation | Powerful but not edge-friendly |
| SVM | N/A | ~5 | ~30 | Light devices | Requires manual features; low memory |
| Random Forest | N/A | ~6 | ~40 | Low-resource devices | Simple; lower accuracy, easy deployment |

Lightweight models like Tiny-LeViT and MobileNetV3, with compact architectures and fast inference, are ideal for resource-constrained contexts. For example, MobileNetV3 maintained a modest computational footprint while achieving an astounding 98.1% accuracy on the Tomato Leaf Disease dataset with just 13 million parameters. Likewise, Tiny-LeViT, a hybrid transformer-CNN model with over 1.1 million parameters, achieved a 97.25% F1-score and a 9% better frame rate on edge devices. Larger transformer-based models, such as PMVT and HvT, on the other hand, had better classification accuracy (99.1% and 99.3%, respectively), but they needed more memory and longer training and inference times, which made them more appropriate for cloud or high-performance computing platforms than edge deployment. This comparison highlights that transformer-based models such as HvT and Swin Transformer provide better accuracy at the expense of larger computing needs, even though MobileNet and Tiny-LeViT are perfect for real-time mobile or embedded application.

### 7.6 Experimental Analysis of CNN and Vision Transformer Models in Terms of Visualization

Plant pest and disease detection relies heavily on visualization, which aids agronomists and researchers in interpreting model predictions, identifying essential characteristics, and increasing the accuracy of disease classification. CNN and Vision Transformer models identify several tomato infections by analyzing vein structure, lesions, color fluctuations, and leaf texture. Recognising the layers that contribute to visualisation improves the explainability of the model and builds confidence in AI-based disease detection. Comparison of attention mechanisms for CNN and Vision Transformer is shown in Table 16.

**Table 16:** Comparison of attention mechanisms

| Feature | CNN (Grad-CAM) | ViT (Attention Maps) |
|---|---|---|
| Attention mode | Spatial feature maps | Global patch-wise attention |
| Localization | Highlights specific disease-affected regions | Distributes attention across multiple patches |
| Accessibility | Provides clear heatmaps | Offers broader context awareness |
| Ideal condition | Detecting localized infections | Identifying complex disease spread patterns |

In this experiment, we use layer-wise visualisation approaches, such as Grad-CAM, Saliency Maps, and Attention Heatmaps, to investigate MobileNetV3, InceptionV3, Hierarchical Vision Transformer (HvT), and Patch-based Multi-view Transformer (PMVT) on the 10-class tomato disease dataset of PlantVillage [34]. The Visualization of disease and pest-affected parts using Attention Map (patches) & Heat Map (red) on the Tomato dataset presented in Fig. 22, where heat maps highlight the severity of the disease by employing red intensity, whereas attention maps pinpoint important areas where the model focuses. This enhances the interpretability of the model and captures contextual dependencies, which aids in accurate diagnosis. Agronomists and farmers can make well-informed decisions and effectively manage plant health with the help of such visual insights. Comparison of Attention Mechanisms for CNNs and Vision Transformer models is shown in Table 16. Additionally, the visualization-related layers and their function of each model used in this experiment for tomato disease detection are presented in Table 17.

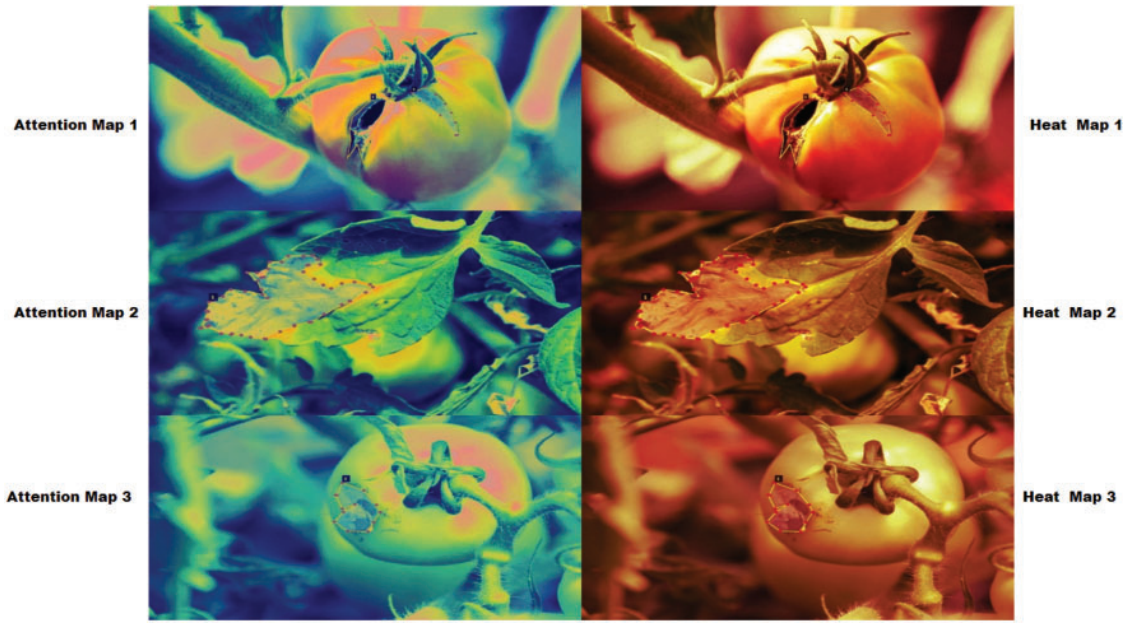

**Figure 22:** The visualization of disease and pest affected parts using attention map (patches) & heat map (red) on tomato dataset

**Table 17:** Layer-wise visualization of models on tomato disease dataset

| Model name | Key visualization layers | Role in visualization |
| --- | --- | --- |
| MobileNetV3 [63] | Last depthwise separable convolution (Conv_dw) | Captures disease-specific leaf textures while ensuring computational efficiency. |
| InceptionV3 [70] | Mixed 7c layer (Final Inception Module) | Enhances lesion detection by merging multi-scale features. |
| HvT (Hierarchical Vision Transformer) [37] | Multi-head self-attention (Stages 3 & 4) | Highlights lesion patterns and fine-grained structures with global context. |
| PMVT (Patch-based Multi-view Transformer) [38] | Final transformer block & class token attention | Analyzes patch-wise features to distinguish disease-specific patterns. |

From the experiment, we got that:

- Transformer models (HvT & PMVT) are useful for complex diseases because they have greater global context awareness.
- Although MobileNetV3 is lightweight, it is less effective in capturing fine-grained information.
- Despite having trouble with class overlap, InceptionV3 performs well by utilizing multi-scale feature extraction.
- Grad-CAM and attention map visualizations validate the role of deep feature layers (Inception and Transformer blocks) in localizing diseased regions.

The interpretability of plant disease categorization is improved by visualization approaches such as Grad-CAM and attention heatmaps, however they have practical limits when used in actual agricultural settings. They may be challenging for non-technical individuals to understand and often rely on high-quality graphics. Furthermore, these visualizations can only be used on edge devices due to the additional computing resources needed to generate them. Future systems should incorporate voice-based instruction, lightweight modules tailored for low-resource contexts, and simplified overlays to improve usability and accessibility. Robustness can also be increased by training models using noisy, real-world images. Effective, farmer-friendly AI-adoption requires striking a balance between explainability and simplicity.

Our investigation shows that Vision Transformers outperform CNNs in terms of visualization efficacy, making them ideal for practical plant disease detection applications.

According to the performance analysis in Table 18, Vision Transformers achieve the highest test accuracy due to their fine-grained feature extraction and strong contextual comprehension.

**Table 18:** Comparison of vision transformers, CNNs, and machine learning models

| Sl. No. | Model type | Model name | Test accuracy (%) | Strengths | Limitations |
|---|---|---|---|---|---|
| 1 | Vision transformer | HvT [37] | 99.3 | Superior global context awareness (via self-attention); excels in fine-grained feature extraction | High training cost and resource demand |
| 2 | Vision Transformer | PMVT [38] | 99.1 | Lightweight, mobile-friendly, high accuracy with fewer parameters | Slight performance drop on mixed symptom datasets |
| 5 | Vision transformer | Tiny-LeViT [39] | 97.8 | Lightweight vision transformer; fast and efficient inference | May underperform on small-scale variation tasks |
| 3 | Vision transformer | ViTAE [40] | 98.9 | Merges CNN priors with attention mechanisms; strong on complex textures | Training complexity; less tested in mobile/real-time setups |
| 4 | Vision transformer | Swin Transformer [36] | 98.2 | Hierarchical attention; well-suited for large datasets | Slower inference speed |
| 6 | CNN | MobileNetV3 [63] | 98.1 | Efficient for edge/mobile deployment; fast training | Slightly lower performance on complex diseases |
| 7 | CNN | InceptionV3 [70] | 93.8 | Multi-scale receptive field; solid baseline CNN | Higher compute cost; less efficient in real-time scenarios |
| 8 | Machine learning | SVM [49] | 90.5 | Performs well on small datasets; effective for binary or few-class problems | Requires handcrafted features; limited generalization |
| 9 | Machine learning | Random Forest [81] | 75.8 | Interpretable and straightforward; handles noisy data | Poor scalability; weak in modeling complex image features |

### 7.7 Environmental Impact of Deep Learning in Agriculture

It is crucial to consider both the advantages and the environmental impacts of deep learning technologies as they are increasingly used in agriculture. Although these models provide practical tools for yield optimization and disease detection, their development and application may have a significant environmental impact. This section examines the sustainability issues surrounding the application of deep learning in agricultural settings.

- **High computational demand:** Deploying deep learning models on a large scale, particularly with transformer-based architectures, can be energy-intensive during the training and inference stages.
- **Carbon footprint concerns:** In the absence of optimization, these models could lead to higher greenhouse gas emissions, which would be in opposition to agricultural sustainability objectives.
- **Efficiency is essential:** Lightweight models, such as Tiny-LeViT or MobileNet, are better suited for locations with limited energy resources and better balance environmental effects and performance.
- **The benefits of edge computing:** It includes deploying models on edge devices (such as smartphones, drones, and Internet of Things sensors) to allow real-time decision-making in the field and lessen reliance on power-hungry centralised servers.
- **Green AI practices:** Strategies like knowledge distillation, model pruning, quantization, and the use of renewable energy sources can considerably decrease the environmental impact.
- **Research direction:** For scalable, environmentally friendly agricultural applications, future research should prioritize sustainable AI architectures and take energy-aware benchmarking into account.

### 7.8 Ethical Considerations in Adopting Deep Learning in Agriculture

Deep learning is becoming a vital component of digital agriculture; therefore, it is essential to consider the ethical implications of its use. These technologies raise questions about data privacy, equitable access, inclusion, and model transparency even as they promise increased productivity and disease management. In particular, in diverse and resource-constrained farming situations, addressing these issues is crucial to guaranteeing that deep learning systems are not only efficient but also equitable, reliable, and available to all stakeholders. Some essential ethical factors are discussed here:

**Data Privacy and Ownership:**

- Sensitive farm image data, which can disclose crop health, location, and agricultural practices, is frequently used by deep learning systems.
- Secure data management procedures, data anonymization, and informed farmer permission are necessary for ethical implementation.
- Data ownership, usage rights, and benefit-sharing from model findings must all be clearly defined by policies.

**The Digital Divide and Equitable Access:**

- In areas with inadequate internet access or limited device availability, the usefulness of AI tools can be restricted.
- Models for low-end devices with offline capabilities and localized interfaces should be created in order to advance equity.
- Governments and non-profit organizations are examples of stakeholders that can make access easier by providing smallholder farmers with training and tool subsidies.

**Inclusivity and Bias:**

- In underrepresented regions or crop kinds, models trained on limited datasets may perform poorly, increasing the risk of inaccurate information or exclusion.
- Diverse, representative training data and field validation in a range of agroclimatic situations are essential for ethical deployment.

**Explainability and Transparency:**

- Models must produce interpretable outputs for AI-based judgments to be reliable and actionable.

- Particularly in crucial situations involving the use of chemicals or the containment of disease, methods like saliency maps or attention visualization should be used to demonstrate which aspects of the image influenced the model's decision.

### 7.9 Enhancing Accessibility and Human-AI Collaboration in Agriculture

Accessibility and inclusivity must coexist with technological innovation if deep learning solutions are to genuinely help smallholder farmers. The adoption of AI-driven solutions in agriculture may be hampered by the linguistic, educational, and digital literacy challenges that many users encounter. Collaboration between humans and AI, user-friendly interfaces, and support for regional languages are essential to address these challenges. To ensure that technology innovation has a practical impact on various farming communities, this section explores methods for making AI technologies more accessible and farmer-friendly.

Smallholder farmers, especially those in developing nations, suffer from linguistic, cultural, and educational obstacles that must be considered to optimize the impact of AI-driven plant disease detection in agriculture. Many farmers may speak regional dialects that are not supported by conventional interfaces, have low literacy rates, or lack technical expertise. Some of the possible solutions discussed here:

**Cross-Linguistic Support:**

- Multilingual interfaces using regional scripts (such as Hindi, Bengali, and Swahili) must be incorporated into mobile apps for accessibility.
- Text-to-speech (TTS) capabilities in regional languages guide users through disease identification and treatment without requiring literacy.

**Simplified User Interfaces:**

- Interfaces should use voice prompts, image-based navigation, and iconography rather than menus with a lot of text.
- Color-coded signs, such as green for healthy and red for unhealthy, are examples of visual cues that can make decision-making easier.
- As seen in PlantVillage Nuru, designing for low-end cellphones with offline capability guarantees inclusivity across different levels of tech access.

**Human-in-the-Loop Design:**

- Artificial intelligence should complement human expertise, not replace it.
  Collaborative workflows may include:
  ◦ Farmers taking images and getting comments.
  ◦ AI outputs are verified and interpreted by local agricultural extension officers.
- Systems can incorporate feedback loops where users validate diagnoses to help models learn and adjust to local circumstances.

**Community-Based Deployment Strategies:**

- Adoption in rural communities can be accelerated by teaching local agri-volunteers or "AI ambassadors" how to use and explain AI tools.
- Farmer-involved co-design workshops yield culturally appropriate solutions and increase technology adoption.

## 8 Conclusion and Research Directions

This review comprehensively analyzed over 150 papers on deep learning-based techniques for identifying plant pests and diseases. We offered a thorough overview of state-of-the-art advancements by structuring the discussion into focused sections, including hyperspectral imaging, visualisation and non-visualisation methods, modified architectures, and transformer models. This framework clarified each approach's strengths, limitations, and gaps in the literature, particularly emphasizing the importance of stage-specific pest and disease detection strategies.

Our review of existing literature made it imperative to benchmark recent advancements against established architectures in plant disease and pest detection. Our experiments demonstrated that transformers (HvT) in computer vision achieved accuracy of 99.3%, as illustrated in Fig. 18. By categorising the study systematically and including the latest papers, we ensure that our review reflects the most current advancements and trends. Our observations offer valuable insights for researchers and practitioners, guiding them in selecting the most suitable techniques for their specific needs and paving the way for future advancements in the field.

To facilitate deployment on low-power edge devices like UAVS, smartphones, and IoT sensors, future research should enhance model efficiency through hardware-aware design, pruning, quantization, and lightweight architectures. Furthermore, developing reliable models trained on heterogeneous, multi-location datasets is necessary to increase generalizability across various agricultural settings, including temperatures, soil types, and plant growth phases. There is potential for improving adaptability and prediction accuracy through domain adaptation, few-shot learning, and multimodal data fusion, such as merging RGB, hyperspectral, and environmental sensor data. Finally, adding explainability mechanisms and developing open-source, scalable frameworks would help end users' trust and real-world adoption.

Our primary objective for this survey is to encourage more researchers to utilise deep learning-based applications in addressing agricultural problems related to plant pests and disease detection. Deep learning algorithms would produce more intelligent, sustainable, and reliable agrarian output.

**Acknowledgement:** The authors would like to thank all contributors and institutions that supported this research.

**Funding Statement:** This research did not receive any specific grant from funding agencies in the public, commercial, or not-for-profit sectors.

**Author Contributions:** MD Tausif Mallick: Led the conception and design of the study, conducted the majority of the literature review, implemented model evaluations, and prepared the initial manuscript draft. He played the primary role in comparing state-of-the-art methods and in interpreting experimental results. Saptarshi Banerjee: Contributed to data analysis, comparative studies of transformer-based models, and supported the survey framework design. He also assisted in model accuracy benchmarking and refining the manuscript's structure. Nityananda Thakur: Focused on statistical evaluation methods and performance metrics for pest and disease detection. He also contributed to organizing and structuring the taxonomy of the techniques. Himadri Nath Saha: Provided overall supervision and critical guidance throughout the manuscript development. He also reviewed and edited all sections for clarity, consistency, and academic rigor. Amlan Chakrabarti: Contributed to the refinement of deep learning methodology discussions and ensured the inclusion of the latest technological insights. He also provided domain-level validation and cross-reviewed key sections of the manuscript. All authors reviewed the results and approved the final version of the manuscript.

**Availability of Data and Materials:** All data and materials used in this study are available from the corresponding author upon reasonable request.

**Ethics Approval:** This study did not involve human or animal participants and therefore did not require ethical approval.

**Conflicts of Interest:** The authors declare no conflicts of interest to report regarding the present study.